\documentclass[letterpaper]{article} 
\usepackage{aaai2026}  
\usepackage{times}  
\usepackage{helvet}  
\usepackage{courier}  
\usepackage[hyphens]{url}  
\usepackage{graphicx} 
\usepackage{natbib}  
\usepackage{caption} 
\usepackage{algorithm}
\usepackage{algorithmic}

\usepackage{amsmath}
\usepackage{booktabs}
\usepackage{amssymb}
\usepackage{subfigure}
\usepackage{newfloat}
\usepackage{listings}
\DeclareCaptionStyle{ruled}{labelfont=normalfont,labelsep=colon,strut=off} 
\floatstyle{ruled}
\newfloat{listing}{tb}{lst}{}
\floatname{listing}{Listing}
\title{SpecDetect: Simple, Fast, and Training-Free Detection of LLM-Generated Text via Spectral Analysis}
\author {
    Haitong Luo\textsuperscript{\rm 1, \rm 2},
    Weiyao Zhang\textsuperscript{\rm 1},
    Suhang Wang\textsuperscript{\rm 3},
    Wenji Zou\textsuperscript{\rm 1, \rm 2},
    Chungang Lin\textsuperscript{\rm 1, \rm 2},\\
    Xuying Meng\textsuperscript{\rm 1, \rm 4$\ast$},
    Yujun Zhang\textsuperscript{\rm 1, \rm 5, \rm 6}\thanks{Corresponding Authors}
}
\affiliations {
    \textsuperscript{\rm 1}Institute of Computing Technology, Chinese Academy of Sciences\\
    \textsuperscript{\rm 2}University of Chinese Academy of Sciences\\
    \textsuperscript{\rm 3}Pennsylvania State University\\
    \textsuperscript{\rm 4}Purple Mountain Laboratory\\
    \textsuperscript{\rm 5}Nanjing Institute of InforSuperBah\\
    \textsuperscript{\rm 6}University of Chinese Academy of Sciences, Nanjing\\
    \{luohaitong21s, mengxuying, nrcyujun\}@ict.ac.cn
}

\usepackage{bibentry}

\begin{document}

\maketitle

\begin{abstract}

The proliferation of high-quality text from Large Language Models (LLMs) demands reliable and efficient detection methods. While existing training-free approaches show promise, they often rely on surface-level statistics and overlook fundamental signal properties of the text generation process. In this work, we reframe detection as a signal processing problem, introducing a novel paradigm that analyzes the sequence of token log-probabilities in the frequency domain. By systematically analyzing the signal's spectral properties using the global Discrete Fourier Transform (DFT) and the local Short-Time Fourier Transform (STFT), we find that human-written text consistently exhibits significantly higher spectral energy. This higher energy reflects the larger-amplitude fluctuations inherent in human writing compared to the suppressed dynamics of LLM-generated text.
Based on this key insight, we construct SpecDetect, a detector built on a single, robust feature from the global DFT: DFT total energy. We also propose an enhanced version, SpecDetect++, which incorporates a sampling discrepancy mechanism to further boost robustness. Extensive experiments show that our approach outperforms the state-of-the-art model while running in nearly half the time. Our work introduces a new, efficient, and interpretable pathway for LLM-generated text detection, showing that classical signal processing techniques offer a surprisingly powerful solution to this modern challenge.

\end{abstract}


\begin{links}
    \link{Code}{https://github.com/luohaitong/SpecDetect}
\end{links}
\section{Introduction}

Recent advancements in Large Language Models (LLMs) have enabled the generation of high-quality text that is often indistinguishable from human writing~\cite{crothers2023machine}. While beneficial, this capability poses significant challenges related to potential misuse, including the spread of misinformation \cite{opdahl2023trustworthy,fang2024bias} and threats to academic integrity \cite{else2023chatgpt,currie2023academic}, which underscore the urgent need for reliable LLM-generated text detection methods.

\begin{figure}[t]
    \centering
    \subfigure[Time-Domain Signal(Log-Probability)]{
        \includegraphics[scale=0.265]{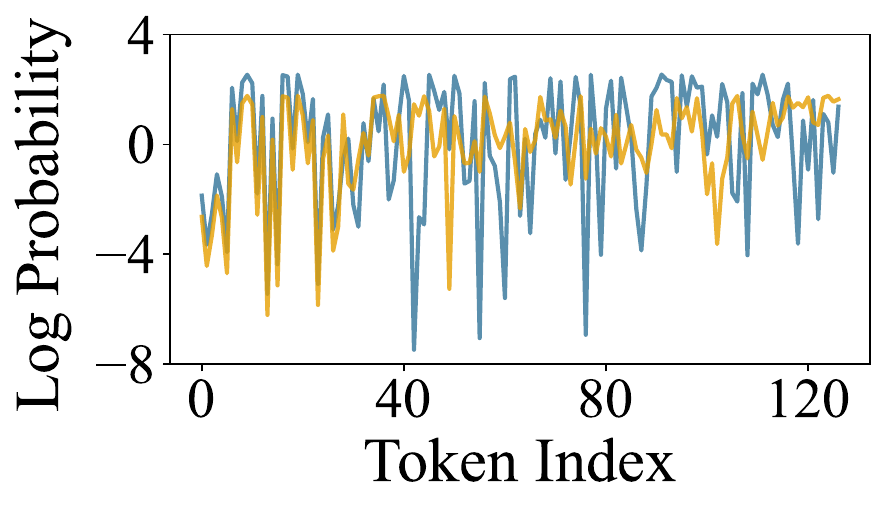}
    }
    \hfill
    \subfigure[Frequency-Domain Spectrum]{
        \includegraphics[scale=0.265]{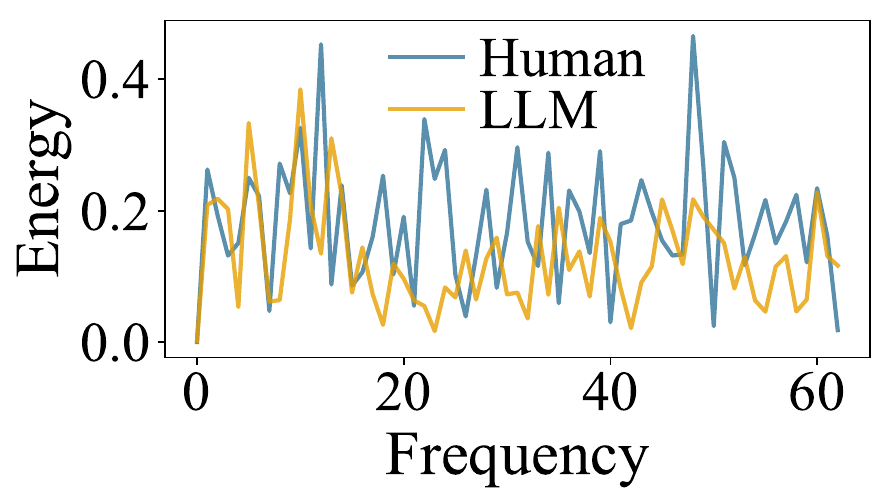}
    }
    \caption{Comparison of a representative human vs. LLM-generated sample (XSum dataset, LLaMA3-8B as source model, GPT-J-6B as proxy model), showing both time-domain signals and the frequency-domain spectrum.}
    \label{fig:qualitative_comparison}
\end{figure}

In contrast, training-free methods offer improved robustness by analyzing intrinsic statistical discrepancies. These approaches implicitly operate on the premise that human writing possesses a greater ``generative vitality'', which is characterized by more dynamic and unpredictable fluctuations in the token probability sequence, than the more constrained output of LLMs. Various methods attempt to model this vitality, evolving from simple global statistics to more sophisticated modeling of the sequence itself, such as treating it as a time series to capture local fluctuations~\cite{xu2024training}. Nevertheless, a key limitation of these time-domain methods is their inability to fully capture the essence of this vitality. This is because the raw, token-by-token probability sequence is often noisy and volatile, making it difficult to distinguish fundamental structural patterns from local, stochastic variations. Consequently, these methods often depend on complex, multi-stage feature engineering pipelines that are sensitive to hyperparameters. This highlights the need for an approach that can effectively capture the subtle differences between human and LLM-generated text in a manner that is both conceptually simple and free from sensitive parameterization.

In this work, we diverge from the time domain to propose a novel \textbf{frequency-domain paradigm} for LLM-generated text detection. We hypothesize that the essential difference between human and machine text lies in the \textbf{dynamic range of its token probability fluctuations}, a property we term ``generative vitality''. We introduce \textbf{SpecDetect}(\textbf{Spec}tral \textbf{Detect}or), a training-free detector that operationalizes this idea by treating the log-probability sequence as a signal and analyzing its spectral properties. The power of a frequency-domain transformation lies in its ability to decompose a signal into its constituent components based on their rate of fluctuation. This process converts the complex time-domain signal into a more structured frequency spectrum, allowing for a precise quantification of its energy and making the hypothesized differences between human and machine text easier to measure. As visualized in Figure~\ref{fig:qualitative_comparison}, this analysis reveals a distinct and consistent separation. The spectrum of human-written text exhibits frequency components with significantly larger magnitudes, which translates to higher overall spectral energy. This provides a direct and robust measure of the greater ``generative vitality'' inherent in human writing.

Our analysis further reveals that a single, hyperparameter-free feature, the DFT Total Energy, serves as a powerful discriminator. This metric effectively quantifies the suppressed ``generative vitality'' of LLM text, which stems from the models' inherent constraint of sampling from high-probability token distributions \cite{holtzman2019curious}, a stark contrast to the unconstrained nature of human expression. Our approach, grounded in classical signal processing theory, is exceptionally simple and computationally efficient. Extensive experiments validate our method, demonstrating that SpecDetect and its sampling-based extension, SpecDetect++, achieve a superior combination of effectiveness and efficiency. Notably, our method outperforms the latest state-of-the-art model with significantly lower computational cost. Our main contributions are:
\begin{itemize}
    \item We are the first to robustly capture the fundamental difference between human and LLM-generated text in the frequency domain, providing a more essential perspective than prior time-domain analyses.
    \item We propose a detector based on a single, hyperparameter-free spectral feature, the DFT Total Energy, that is simpler and faster than current state-of-the-art methods.
    \item We empirically demonstrate our method sets a new state-of-the-art in the effectiveness-efficiency trade-off, outperforming the prior SOTA at half the runtime.
\end{itemize}

\section{Related Work}
\label{sec:related_work}

\subsection{Detection of LLM-Generated Text}
Existing detection approaches are broadly categorized into training-based and training-free methods. Training-based detectors learn a classifier \cite{bhattacharjee2023conda,li2023deepfake,tian2023multiscale} but often suffer from poor generalization to out-of-distribution data \cite{uchendu2020authorship,chakraborty2023possibilities}. To overcome this, training-free methods identify intrinsic statistical differences between human and machine text. These methods have evolved significantly, beginning with simple global statistics like average Log-Likelihood and Log-Rank \cite{solaiman2019release}. A major advancement is the distributional discrepancy paradigm, pioneered by DetectGPT \cite{mitchell2023detectgpt}, which contrasts original text with its perturbed versions, though often at a high computational cost. Subsequent work like Fast-DetectGPT \cite{bao2023fast} has aimed to improve the efficiency of this approach. More recently, the state-of-the-art method, Lastde \cite{xu2024training}, introduces a time-series perspective to analyze the token probability sequence. However, its innovative approach relies on a complex, multi-stage feature engineering pipeline with sensitive hyperparameters, suggesting that a simpler, more fundamental signal has yet to be found. Our work diverges from this time-domain analysis, proposing that this essential distinction lies in the frequency domain.

\subsection{Frequency-Domain Methods in NLP}
Frequency-domain analysis is a cornerstone of signal processing and has been applied to NLP tasks like sentiment analysis \cite{chakraborty2022aspect} and to efficiently approximate self-attention in Transformers \cite{lee2021fnet}. However, to our knowledge, no prior work has applied spectral analysis to the token probability sequence for LLM text detection. Our work is the first to bridge this gap, demonstrating that a direct frequency-domain analysis can reveal fundamental and highly discriminative features.

\section{The SpecDetect Method}
\label{sec:method}

\begin{figure*}[t]
    \centering
    \includegraphics[scale=0.48]{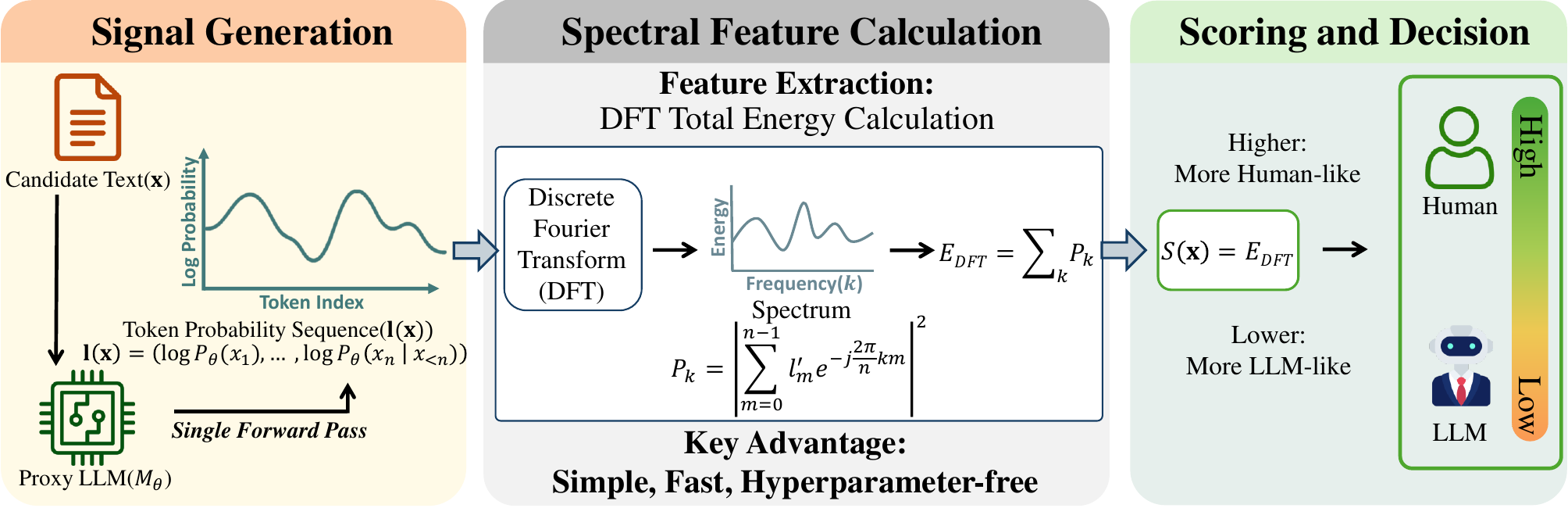}
    \caption{The overall framework of the SpecDetect method. The process consists of three main stages: (1) Inference to obtain the token probability sequence, (2) Frequency-domain transformation to compute the DFT Total Energy, and (3) Scoring to make a final classification.}
    \label{fig:framework}
\end{figure*}

In this section, we introduce our proposed method, SpecDetect. Our approach is grounded in a frequency-domain analysis of the token probability sequence, from which we derive and systematically evaluate a suite of candidate spectral features. Based on this investigation, which identifies DFT Total Energy as the most robust indicator, we formally define two detectors: the simple, single-feature SpecDetect and its sampling-based extension, SpecDetect++. The overall pipeline of our method is illustrated in Figure~\ref{fig:framework}.

\subsection{Frequency-Domain Analysis}
\label{subsec:analysis}

\subsubsection{A Signal Processing Perspective.}

Our analysis begins with the core intuition that human and machine writing differ in ``generative vitality." Human text, with its surprising word choices, produces high-amplitude fluctuations in its token probability sequence. In contrast, LLMs are fundamentally constrained to a high-probability vocabulary \cite{holtzman2019curious}, resulting in lower-amplitude variations. While present in the time domain, these amplitude differences are obscured by the signal's noisy nature. We hypothesize that the frequency domain provides a more effective means to quantify these fundamental fluctuation patterns.

To test this, we adopt a novel perspective: we treat the token probability sequence as a signal. Given an input text $\mathbf{x} = (x_1, \dots, x_n)$ and a proxy LLM $M_\theta$, we first obtain the token log-probability sequence $\mathbf{l}(\mathbf{x}) = (l_0, l_1, \dots, l_{n-1})$. Each element $l_i$ is the log-probability of the $(i+1)$-th token $x_{i+1}$, which is calculated as:
\begin{equation}
    l_i = \log P_\theta(x_{i+1} \mid x_{<i+1}).
\end{equation}
Here, $P_{\theta}$ is the probability distribution of the language model parameterized by $\theta$, and $x_{<i+1}$ represents the preceding context tokens $(x_1, \dots, x_{i})$. We regard this sequence $\mathbf{l}(\mathbf{x})$ as a \textbf{discrete-time signal}. To focus on its dynamic fluctuations, we make the signal zero-mean: $\mathbf{l}'(\mathbf{x}) = \mathbf{l}(\mathbf{x}) - \mu_{\mathbf{l}}$, where $\mu_{\mathbf{l}}$ is the mean of the sequence $\mathbf{l}(\mathbf{x})$.

\subsubsection{Spectral Feature Extraction.}



We hypothesize a key difference between human and machine-generated text lies in the dynamic fluctuations of their token log-probabilities. To characterize these (e.g., amplitude and speed), we analyze the log-probability sequence as a frequency-domain signal, moving beyond individual token probabilities to examine its overall structure. We adopt two complementary tools: the Discrete Fourier Transform (DFT)~\cite{holtzman2019curious} for a global view and the Short-Time Fourier Transform (STFT) \cite{griffin1984signal} for a local perspective.

The DFT provides a global view by decomposing the zero-mean log-probability sequence $\mathbf{l}'(\mathbf{x})$ into constituent frequency components, revealing dominant frequencies across the text. For a real-valued input, it produces a symmetric sequence of complex numbers $\mathbf{X} = (X_0, \dots, X_{n-1})$, where:
\begin{equation}
    X_k = \mathcal{F}(\mathbf{l}')_k = \sum_{m=0}^{n-1} l'_m e^{-j \frac{2\pi}{n} km},
\end{equation}
with $\mathcal{F}$ as the DFT operator and $k$ as the frequency index. $X_k$ is a complex value and $P_k = |X_k|^2$ denotes energy at frequency $k$. We focus on the first $n/2$ components since they contain all the unique frequency information.

In contrast, the STFT offers a local perspective by computing DFT over short, overlapping windows, revealing how frequency content evolves from beginning to end. Formally:
\begin{equation}
    S(f, t) = \sum_{m=0}^{L-1} l'_{t+m} w_m e^{-j \frac{2\pi}{L} fm},
\end{equation}
where $f$ is the frequency index, $t$ is the token position, and $w_m$ is a window function. Here we use a Hann window \cite{shimauchi2014accurate} of length $L=20$ and hop size $h=10$. $|S(f, t)|^2$ represents the energy of each frequency at each time window.



\begin{figure*}[t]
    \centering
    \includegraphics[width=1\linewidth]{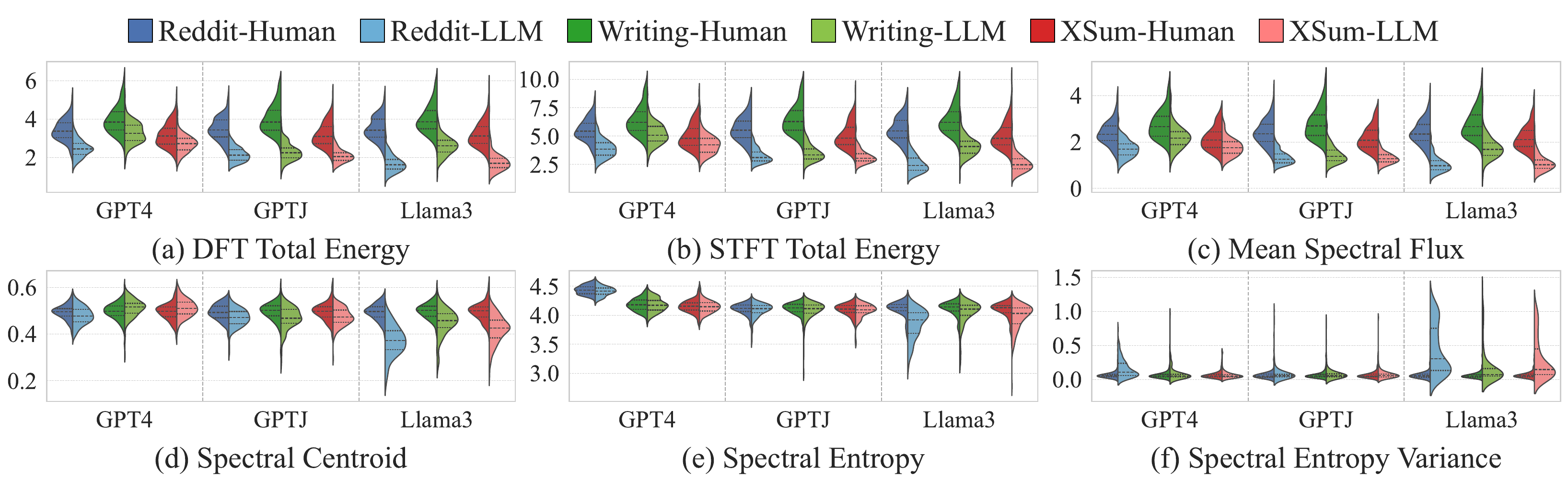}

    \caption{Distributions of spectral features for human and LLM-generated text across different datasets and source models. The top row displays energy-based features ($E_{DFT}$, $E_{STFT}$, $\overline{F}_{spec}$), while the bottom row shows frequency-based features ($C_{spec}$, $H_{spec}$, $V_{H_{spec}}$). The visualization clearly demonstrates that energy-based features provide a strong and consistent separation between the two classes, whereas the distributions for frequency-based features show significant overlap.}
    \label{fig:spectral_distributions}
\end{figure*}


From the spectral transformations, we extract two feature categories: energy-based features and frequency-based features. Specifically, as energy directly reflects the magnitude of sequence fluctuations, the energy-based features below quantify the amplitude of log-probability fluctuations:
\begin{itemize}
    \item \textbf{DFT Total Energy} ($E_{DFT}$): Measures total fluctuation intensity by summing energy across all frequency components: $E_{DFT} = \sum_{k=0}^{n/2} P_k$. A higher value indicates greater overall variation.
    \item \textbf{STFT Total Energy} ($E_{STFT}$): Measures total local fluctuation intensity by aggregating spectral energy across all time windows and frequencies: $E_{STFT} = \sum_{t=0}^{T-1} \sum_{f=0}^{L-1} |S(f, t)|^2$. A higher value suggests larger cumulative local fluctuations.
    \item \textbf{Mean Spectral Flux} ($\overline{F}_{spec}$): Captures energy ``fluidity" via the change in local spectra: $\overline{F}_{spec} = \frac{1}{T-1} \sum_{t=1}^{T-1} \sum_f |{S}(f, t) - {S}(f, t-1)|$. A lower value indicates more stable, self-similar signals over time.
\end{itemize}

Frequency-based features, by contrast, characterize the shape and complexity of the spectrum, which correspond to the speed and dynamics of the signal's fluctuations:
\begin{itemize}
    \item \textbf{Spectral Centroid} ($C_{spec}$): Identifies the spectrum's ``center of mass" (average fluctuation speed): $C_{spec} = \frac{\sum_{k=0}^{n/2} k \cdot P_k}{\sum_{k=0}^{n/2} P_k}$. Higher values indicate faster variations.
    \item \textbf{Spectral Entropy} ($H_{spec}$): Measures global spectrum uniformity: $H_{spec} = -\sum_{k=0}^{n/2} {P}_k \log_{2}({P}_k)$. A lower value signifies energy concentration in fewer frequencies (more predictable/less complex signals).
    \item \textbf{Spectral Entropy Variance} ($V_{H_{spec}}$): Measures stability of complexity over time (variance of local spectral entropy): $V_{H_{spec}} = \text{Var}_t[H_{spec}(t)]$ where $H_{spec}(t) = -\sum_{f=0}^{L-1} {S}(f, t) \log_{2}({S}(f, t))$. Lower values indicate more consistent frequency distribution across windows.
\end{itemize}
These spectral metrics enable comprehensive quantitative analysis of human and LLM-generated texts.

\subsubsection{Analysis and Selection of Spectral Features.}
To systematically explore spectral differences between human and LLM-generated text, we analyze three datasets: XSum \cite{narayan2018don}, Reddit ELI5 \cite{fan2019eli5}, and WritingPrompts \cite{fan2018hierarchical}. Using source models GPT-J \cite{wang2023seqxgpt}, LLaMA3-8B \cite{llama3modelcard}, and GPT-4-Turbo \cite{achiam2023gpt} (with GPT-J as proxy), we test both white-box (GPT-J) and black-box (LLaMA3-8B, GPT-4-Turbo) conditions. Figure~\ref{fig:qualitative_comparison} shows a sample pair for intuition (STFT visualizations in Appendix Figure~\ref{fig:stft_heatmap}), suggesting human text has a more volatile time-domain signal, with a spectrum of significantly higher overall energy.

To verify this observation statistically, we evaluate our candidate features by visualizing their distributions in Figure~\ref{fig:spectral_distributions}. The results reveal a decisive pattern. The frequency-based metrics, which characterize the speed of fluctuations, show substantial overlap between human and LLM text. This suggests that due to sampling strategies like top-k, LLMs can produce variations in token choice at a speed that is not distinctly different from that of human writing.

However, the \textbf{energy-based metrics consistently and significantly separate the two classes}. This confirms our core hypothesis. Because LLM decoding is largely restricted to a narrow set of high-probability tokens \cite{holtzman2019curious}, the \textbf{amplitude} of its log-probability fluctuations is inherently constrained and much smaller than the wider, more varied fluctuations found in human text. This low-amplitude signature is a fundamental artifact of the generation process.


Given that the energy-based metrics are most effective, we perform a Pearson correlation analysis to understand their relationships. The analysis reveals that \textbf{all three metrics are highly correlated with one another}: the correlation between the global $E_{DFT}$ and the local $E_{STFT}$ is exceptionally high (0.97), and both are also strongly correlated with the Mean Spectral Flux, $\overline{F}_{spec}$ (0.89 and 0.94, respectively). This strong linear relationship suggests that all three features capture the same underlying low-amplitude phenomenon and are therefore largely redundant. Given this informational redundancy, and in accordance with the principle of parsimony, we select the \textbf{DFT Total Energy} ($E_{DFT}$) as our single robust feature, as it is a global, hyperparameter-free, and the simplest of the three metrics.

\begin{table*}
\centering
\resizebox{\linewidth}{!}{
\begin{tabular}{lccccccccccccc}
\toprule
\textbf{Method} & \textbf{GPT-2} & \textbf{Neo-2.7} & \textbf{OPT-2.7} & \textbf{LLaMA-13} & \textbf{LLaMA2-13} & \textbf{LLaMA3-8} & \textbf{OPT-13} & \textbf{BLOOM-7.1} & \textbf{Phi-4} & \textbf{Qwen3-8} & \textbf{Claude-3} & \textbf{GPT4-Turbo}  & \textbf{Avg.}\\
\midrule
\multicolumn{14}{c}{\textit{sample-based}} \\ \hline
Likelihood       & 0.6645 & 0.6710 & 0.6740 & 0.6575 & 0.6861 & 0.9730 & 0.6880 & 0.6180 & 0.7185 & \underline{0.9838} & 0.9838 & \underline{0.7970}  & 0.7596 \\
LogRank & 0.7076 & 0.7116 & 0.7228 & 0.7030 & 0.7264 & \underline{0.9789} & 0.7297 & 0.6749 & \underline{0.7756} & \textbf{0.9871} & 0.2104 & 0.7920  & 0.7267 \\
Entropy & 0.6123 & 0.5866 & 0.5454 & 0.4913 & 0.4517 & 0.2028 & 0.5308 & 0.6084 & 0.3910 & 0.2840 & \underline{0.9858} & 0.3510  & 0.5034 \\
DetectLRR & 0.7916 & 0.7921 & 0.8188 & 0.7850 & 0.7882 & 0.9603 & 0.8013 & 0.7955 & 0.7662 & 0.9710 & 0.9777 & 0.7384  & 0.8322 \\
Lastde & \underline{0.9011} & \underline{0.9040} & \underline{0.8989} & \textbf{0.8068} & \underline{0.8019} & \textbf{0.9851} & \underline{0.8990} & \textbf{0.8834} & 0.7353 & 0.9654 & \textbf{0.9893} & 0.7609  & \underline{0.8776} \\
\textbf{SpecDetect} & \textbf{0.9078} & \textbf{0.9171} & \textbf{0.9118} & \underline{0.8051} & \textbf{0.8090} & 0.9761 & \textbf{0.9005} & \underline{0.8781} & \textbf{0.7773} & 0.9783 & 0.9833 & \textbf{0.8054}  & \textbf{0.8875} \\ \hline
\multicolumn{14}{c}{\textit{distribution-based}} \\ \hline
DetectGPT & 0.6933 & 0.7683 & 0.6775 & 0.6483 &0.6925 &0.7550 &0.7342 &0.5767 &0.7733 &0.8275 &0.9308 &0.7525 &0.7358 \\
DetectNPR &0.7133 &0.7683 &0.6817 &0.6808 &0.7167 &0.9450 &0.7300 &0.6367 &0.7575 &0.9383 &0.9542 &0.7758 &0.7749 \\
DNA-GPT &0.6254 &0.6295 &0.6362 &0.6029 &0.6649 &0.9581 &0.6777 &0.6254 &0.7250 &0.9558 &0.9592 &0.6860 &0.7288\\
Fast-DetectGPT & 0.8967 & 0.8874 & 0.8651 & 0.7758 & 0.7761 & \underline{0.9724} & 0.8616 & 0.8458 & 0.7992 & \textbf{0.9877} & \textbf{0.9996} & \underline{0.8818}  & 0.8791 \\
Lastde++ & \underline{0.9463} & \underline{0.9543} & \underline{0.9395} & \underline{0.8517} & \underline{0.8556} & \textbf{0.9842} & \underline{0.9350} & \underline{0.9215} & \textbf{0.8534} & 0.9720 & \underline{0.9994} & \underline{0.8818}  & \underline{0.9246} \\
\textbf{SpecDetect++} & \textbf{0.9518} & \textbf{0.9587} & \textbf{0.9465} & \textbf{0.8636} & \textbf{0.8643} & 0.9686 & \textbf{0.9360} & \textbf{0.9333} & \underline{0.8289} & \underline{0.9867} & 0.9896 & \textbf{0.8833} & \textbf{0.9259} \\
\bottomrule
\end{tabular}
} 
\caption{Detection AUC under black box scenarios, with values averaged across three datasets: XSum, WritingPrompts, and Reddit. The last column (``Avg") denotes the mean AUC across all source models, calculated over the three datasets. Best and second-best results are highlighted in \textbf{bold} and \underline{underline}, respectively.}
\label{tab:blackbox_results}
\end{table*}

\subsection{The Proposed Detector: SpecDetect}
\label{subsec:specdetect}

Based on our frequency-domain analysis, we propose SpecDetect, a simple and training-free detector. As shown in Figure \ref{fig:framework}, the detection process follows three main steps:
\begin{enumerate}
    \item \textbf{Signal Generation:} For a given candidate text $\mathbf{x}$, retrieve its zero-mean log-probability sequence, $\mathbf{l}'(\mathbf{x})$, using a proxy model $M_\theta$.
    
    \item \textbf{Spectral Feature Calculation:} Apply the DFT to the signal to convert it to the frequency domain and compute its DFT total energy, $E_{DFT} = \sum_{k=0}^{n/2} |X_k|^2$, where $X_k$ is the $k$-th component of the DFT output.
    
    \item \textbf{Scoring and Descision:} Use the DFT total energy $E_{DFT}$ as the final detection score. A higher score indicates a greater likelihood that the text is human-written.
\end{enumerate}
The entire process of deriving the SpecDetect score $S(\mathbf{x})$ from $\mathbf{l}(\mathbf{x})$ can be concisely summarized as:
\begin{equation}
    S(\mathbf{x}) = \sum_{k=0}^{n/2} | \mathcal{F}(\mathbf{l}(\mathbf{x}) - \mu_{\mathbf{l}})_k |^2
\end{equation}
where $\mu_{\mathbf{l}}$ is the mean of the sequence $\mathbf{l}(\mathbf{x})$, and $\mathcal{F}$ denotes the Discrete Fourier Transform operator. Following prior work \cite{bao2023fast,xu2024training}, we treat the output score $S(\mathbf{x})$ as the likelihood of text being human-written. For a sequence of length $n$, the naive DFT has a time complexity of $O(n^2)$, but its efficient implementation via the Fast Fourier Transform (FFT) \cite{cooley1965algorithm} reduces this to $O(n \log n)$, ensuring computational efficiency.

\subsection{Extension with Sampling: SpecDetect++}
\label{subsec:specdetect++}


To further enhance robustness, we introduce \textbf{SpecDetect++}, including a sampling discrepancy method similar to previous work \cite{bao2023fast}. The core intuition is that LLM text scores are typical within their LLM-generated variation distribution, while human text scores act as outliers.

To formalize this, for a given text $\mathbf{x}$ and a proxy model $M_\theta$, we first consider a distribution of contrastive samples $\{\tilde{\mathbf{x}}\}$. Following \cite{xu2024training}, we generate these samples using the same model, such that $\tilde{\mathbf{x}} \sim M_{\theta}(\cdot|\mathbf{x})$. We then compute the mean $\mu_S$ and variance $\sigma_S^2$ of the base SpecDetect scores, $S(\tilde{\mathbf{x}})$, over this sample distribution. Formally, these statistics are the expectations:
\[
\mu_{S} = \mathbb{E}_{\tilde{\mathbf{x}} \sim M_{\theta}} [ S(\tilde{\mathbf{x}}) ] \quad \text{and} \quad \sigma_{S}^2 = \mathbb{E}_{\tilde{\mathbf{x}} \sim M_{\theta}} [ ( S(\tilde{\mathbf{x}}) - \mu_{S} )^2 ].
\]
With the mean and standard deviation of the sample distribution, the final SpecDetect++ score for the original text, $S_{++}(\mathbf{x})$, is calculated as its z-score:
\[
S_{++}(\mathbf{x}) = \frac{S(\mathbf{x}) - \mu_{S}}{\sigma_{S}},
\]
where $S(\mathbf{x})$ is the base SpecDetect score of the original text. In practice, $\mu_S$ and $\sigma_S$ are estimated by generating a finite number of samples $N$. This normalization sharpens the classification boundary and improves robustness against adversarial attacks by amplifying the intrinsic differences between human and machine-generated text.

\section{Experiments}
\label{sec:experiments}

\subsection{Experimental Setup}
\label{subsec:setup}

\paragraph{Datasets.}
We evaluate our methods on a diverse suite of datasets, following the configurations of prior work \cite{bao2023fast,xu2024training}. Our experiments are conducted in two settings: a white-box scenario using the XSum \cite{narayan2018don}, SQuAD \cite{rajpurkar2016squad}, and WritingPrompts \cite{fan2018hierarchical} datasets, and a more realistic black-box scenario using XSum, Reddit ELI5 \cite{fan2019eli5}, and WritingPrompts. To assess performance on other languages, our analysis also includes two non-English datasets: WMT16-De (German) \cite{bojar2016findings} and Zhihu-Economy (Chinese) \cite{xu2024training}. In all cases, LLM-generated text is created by prompting source models with the initial 30 tokens of each human-written example, ensuring the human and machine-generated texts have matching lengths. More details are in Appendix \ref{app:dataset}.

\paragraph{Models.}

We evaluate our methods under two distinct scenarios: white-box and black-box. Following previous work \cite{xu2024training}, the white-box setting assumes that the proxy model used for detection is the same as the source model that generated the text.
In contrast, the more challenging and realistic black-box setting assumes that the detector does not have access to the source model. Therefore, a different proxy model must be used. For each setting, we test against a diverse set of 12 source models, encompassing a wide range of open-source architectures and state-of-the-art closed-source models. Unless otherwise specified, the open-source GPT-J-6B serves as the sole proxy model for all black-box tasks. Detailed specifications for all models are provided in Appendix \ref{app:imple}.



\paragraph{Baselines.}

We compare our approach against 10 training-free baselines, which are divided into two groups (see Appendix \ref{app:baselines} for full details). Sample-based methods directly use token statistics for scoring and include Log-Likelihood \cite{solaiman2019release}, LogRank \cite{solaiman2019release}, Entropy \cite{gehrmann2019gltr,ippolito2019automatic}, DetectLRR \cite{su2023detectllm}, and Lastde \cite{xu2024training}. In contrast, distribution-based methods score text by contrasting it with generated samples and include DetectGPT \cite{mitchell2023detectgpt}, DetectNPR \cite{su2023detectllm}, DNA-GPT \cite{yang2023dna}, Fast-DetectGPT \cite{bao2023fast}, and Lastde++ \cite{xu2024training}. Consistent with prior work \cite{xu2024training}, we use AUC as the evaluation metric and set the sampling number for SpecDetect++ to 100.

\subsection{Effectiveness-Efficiency Trade-Off}
\subsubsection{Detection Performance.}
\label{subsec:detection_performance}





We primarily evaluate our methods in the black-box setting, which most closely mirrors real-world applications, though comprehensive white-box results are also provided in Appendix \ref{app:white_box} for completeness. Table \ref{tab:blackbox_results} presents the average detection AUC across three datasets, while the detailed per-dataset results can be found in Appendix \ref{app:black_box}. Our analysis demonstrates that the SpecDetect framework establishes a new state-of-the-art in both categories of training-free detection.

Among single-pass, sample-based methods, our base detector, SpecDetect, consistently outperforms the previous SOTA, Lastde. As shown in Table \ref{tab:blackbox_results}, SpecDetect achieves a higher average AUC (0.8875 vs. 0.8776) and shows particular strength against a wide range of architectures, including both older models (e.g., GPT-2, Neo-2.7) and modern closed-source models like GPT-4 Turbo. This highlights the broad generalizability of our spectral energy feature.

In the more powerful distribution-based category, our enhanced method, SpecDetect++, also achieves state-of-the-art performance. It surpasses the prior SOTA, Lastde++, with a higher average AUC (0.9259 vs. 0.9246) and demonstrates more consistent, robust detection across the majority of tested models. This is a critical advantage for real-world scenarios where the source model is unknown. In summary, SpecDetect and SpecDetect++ each set a new state-of-the-art in their respective classes, offering an unparalleled combination of performance for training-free detection.

\subsubsection{Efficiency Analysis.}
\label{subsec:efficiency}

Beyond detection accuracy, computational efficiency is a critical requirement for any practical detector. To evaluate this, we measure the average inference time per sample for each method under a specific black-box condition: detecting text generated by GPT-4-Turbo on the XSum dataset, using GPT-J-6B as the proxy model. All runtimes were measured on a single NVIDIA H800 GPU.

\begin{figure}[t]
    \centering
    \includegraphics[width=1.\linewidth]{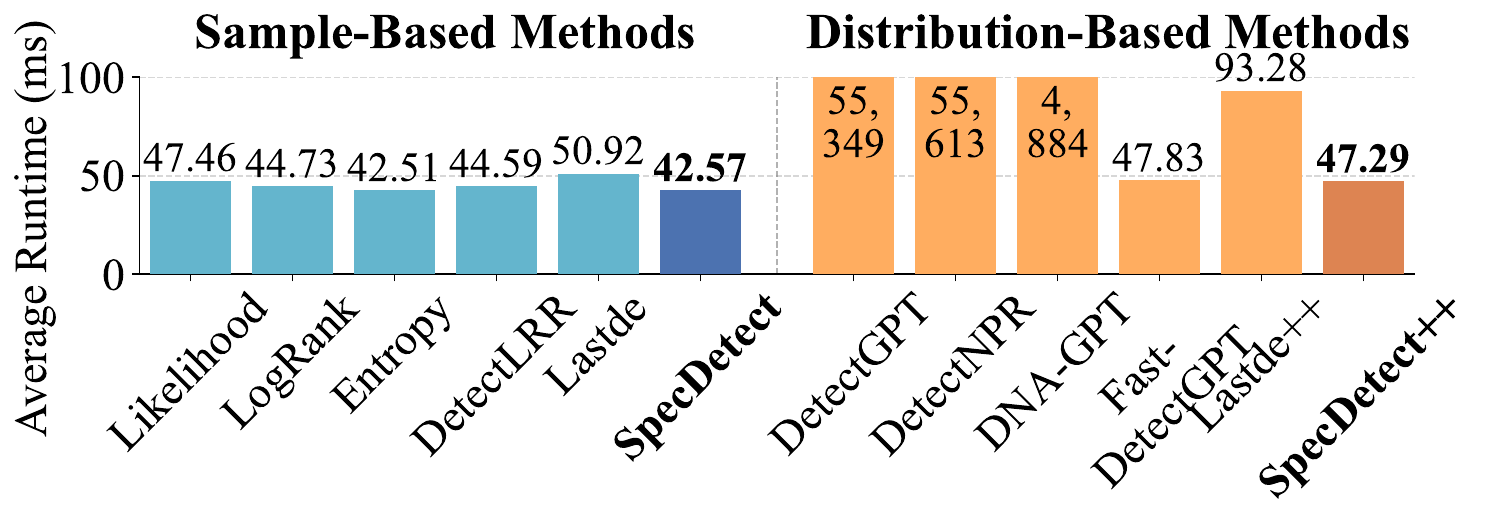}
    \caption{Efficiency comparison of detection methods. Runtimes are averaged per sample (in milliseconds) and labeled on each bar. The bars for perturbation-based methods with extremely long runtimes have been truncated.}
    \label{fig:efficiency}
\end{figure}

The results, visualized in Figure~\ref{fig:efficiency}, clearly show that our methods achieve high performance at a significantly lower computational cost. Specifically, our base method SpecDetect is \textbf{16.4\% faster} than its main competitor, Lastde. More strikingly, our enhanced method, SpecDetect++, is nearly \textbf{twice as fast} as the SOTA model Lastde++. The results show that our method achieves superior performance at approximately \textbf{half the computational cost} of the latest SOTA.

\subsubsection{Trade-off Comparison.}
\begin{figure}[t]
    \centering
    \includegraphics[width=1\linewidth]{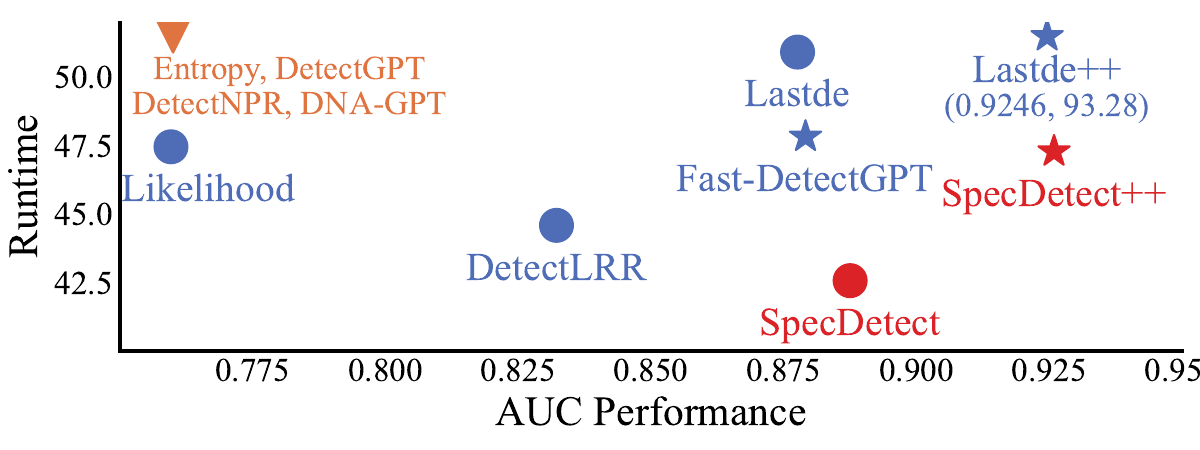} 
    \caption{Performance vs. Efficiency trade-off. The x-axis represents detection performance (AUC), and the y-axis represents average runtime (in milliseconds). Methods are grouped by category using marker shapes, and our proposed methods are highlighted in red. Several methods (i.e., Entropy, DetectGPT, DetectNPR, DNA-GPT) are excluded as outliers due to non-competitive performance or prohibitive runtimes. The coordinates for Lastde++ are explicitly labeled because its runtime exceeds the y-axis range.}
    \label{fig:tradeoff_plot}
\end{figure}

A critical aspect of a practical detector is its ability to balance high performance with computational efficiency. To visualize this trade-off, we plot the average black-box detection AUC against the average inference runtime for each method, with the results presented in Figure~\ref{fig:tradeoff_plot}. The ideal detector is located in the bottom-right corner, representing both high accuracy and low runtime.

Our methods demonstrate a superior performance-efficiency profile, with SpecDetect achieving the optimal balance within the fast, sample-based category by having the lowest runtime and a high AUC. It's notably effective, outperforming more complex methods like Fast-DetectGPT despite being significantly faster. For the high-performance, distribution-based category, our enhanced method, SpecDetect++, sets a new state-of-the-art, surpassing the previous best, Lastde++, with a higher AUC at approximately half the computational cost. In summary, our SpecDetect framework dominates the performance-efficiency landscape, offering an unparalleled combination of speed and effectiveness for training-free detectors.

\subsection{Robustness and In-depth Analysis}
\label{subsec:robustness}

We evaluate our method's robustness in challenging, black-box scenarios. Further analyses on decoding strategies and alternative detectors are in Appendix \ref{app:decoding} and \ref{app:alter_metrics}.

\paragraph{Robustness to Paraphrasing Attacks.}
\begin{figure}[t]
    \centering
    \includegraphics[width=1.\linewidth]{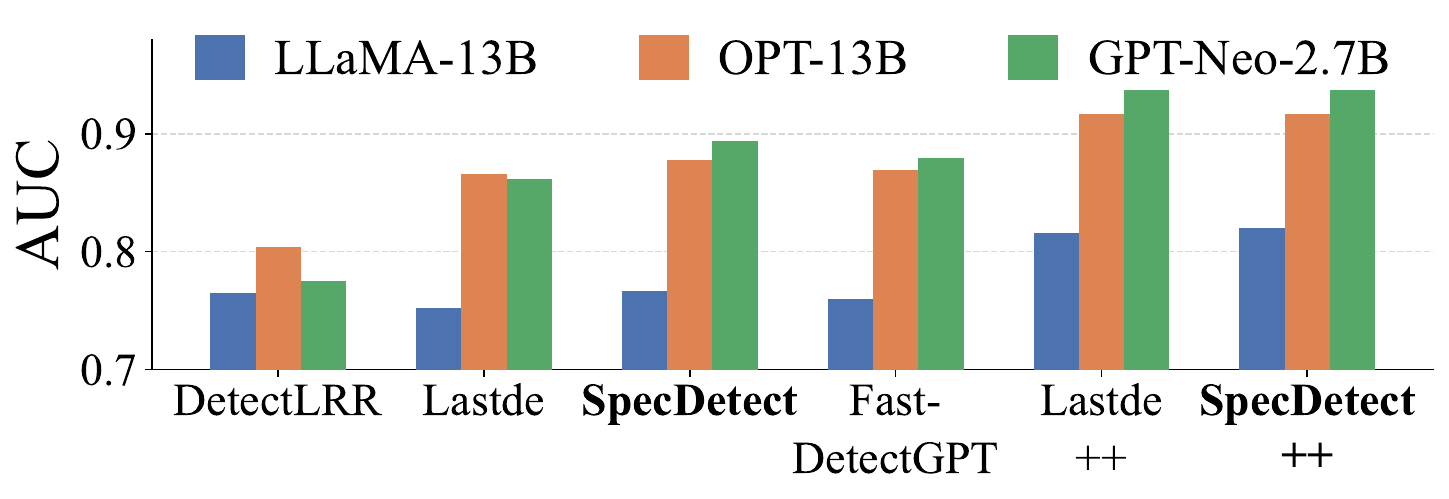}
    \caption{Average AUC performance on paraphrased texts under a black-box setting. Each group of bars represents a different source model.}
    \label{fig:paraphrase_results}
\end{figure}

Paraphrasing attacks rewrite human and LLM-generated texts to alter style while preserving meaning, significantly degrading detector performance \cite{krishna2023paraphrasing, sadasivan2023can}. Following previous work \cite{bao2023fast,xu2024training}, we evaluate this Robustness by using a T5-Paraphraser \cite{alisetti2021paraphrase} on texts from three source models: GPT-Neo-2.7B, LLaMA-13B, and OPT-13B. Figure~\ref{fig:paraphrase_results} shows average results across three datasets (details in Appendix Table \ref{app_tab:para_attack}).



Results highlight the robustness of our frequency-domain approach. SpecDetect++ is highly resilient, matching state-of-the-art Lastde++ under adversarial conditions. Notably, our base method SpecDetect consistently outperforms baselines—critical evidence that its core spectral energy feature inherently resists paraphrasing-induced changes. SpecDetect’s strong performance without sampling indicates the underlying spectral signature is a fundamental property of generated text, ensuring robustness against such attacks.

\paragraph{Impact of Text Length.}
Prior studies show shorter texts are harder to detect \cite{verma2023ghostbuster, mao2024raidar}. We investigate this using texts generated by the LLaMA-13B source model across the XSum, WritingPrompts, and Reddit datasets. Since the first 30 tokens of each human-LLM pair are identical, we evaluate on texts truncated to lengths of 60, 90, 120, and 150 words. Figure~\ref{fig:text_length} shows the average AUC across these three datasets by text length, with detailed results in Appendix \ref{app:text_length}. Consistent with prior work, most methods improve with longer texts. Importantly, our SpecDetect and SpecDetect++ consistently outperform competitors across all lengths, with the gap widening for longer texts. This highlights that our spectral energy feature becomes increasingly powerful with signal length, underscoring its fundamental robustness.



\begin{figure}[t]
    \centering
    \subfigure[Impact of Text Length]{
        \includegraphics[scale=0.1818]{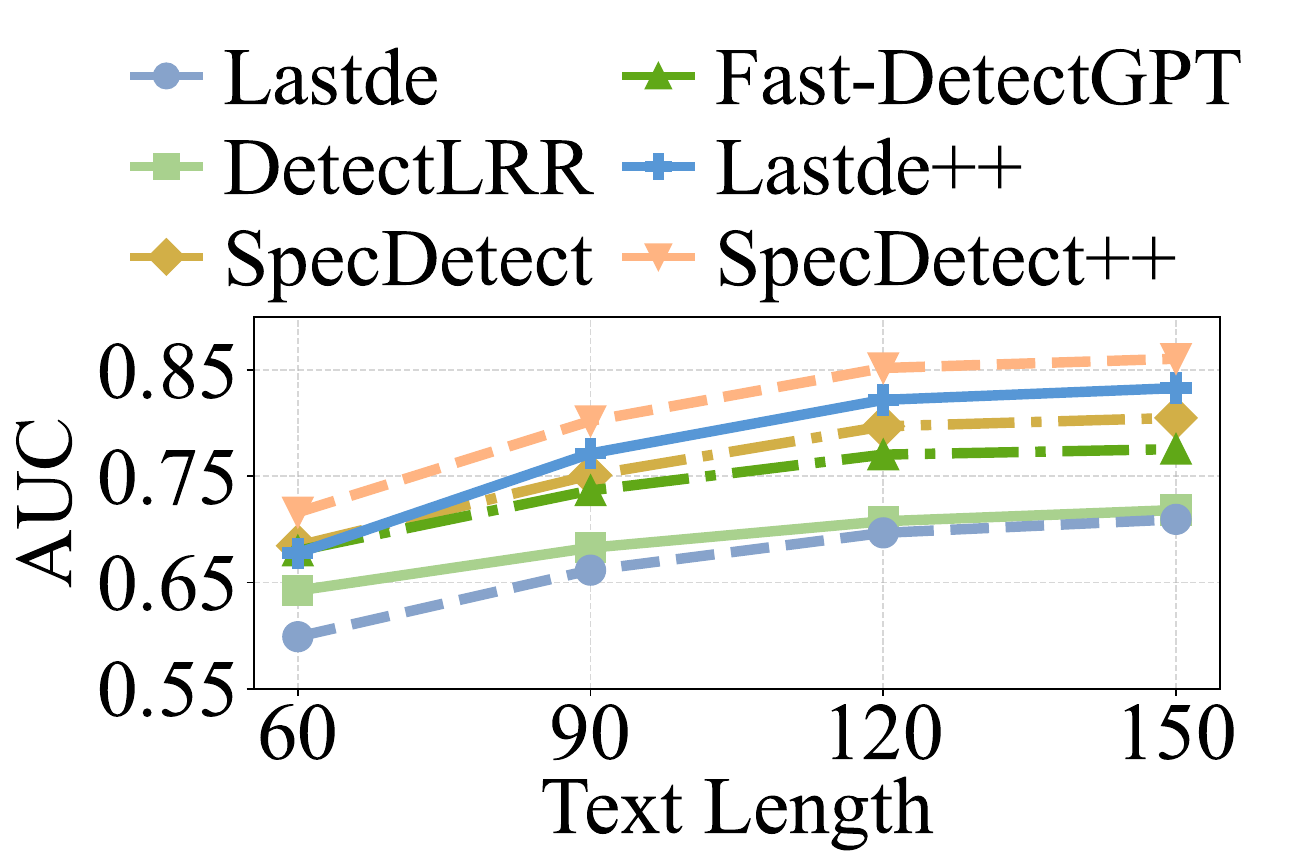}
        \label{fig:text_length}
    }
    \hfill
    \subfigure[Impact of Contrast Sample Size]{
        \includegraphics[scale=0.1818]{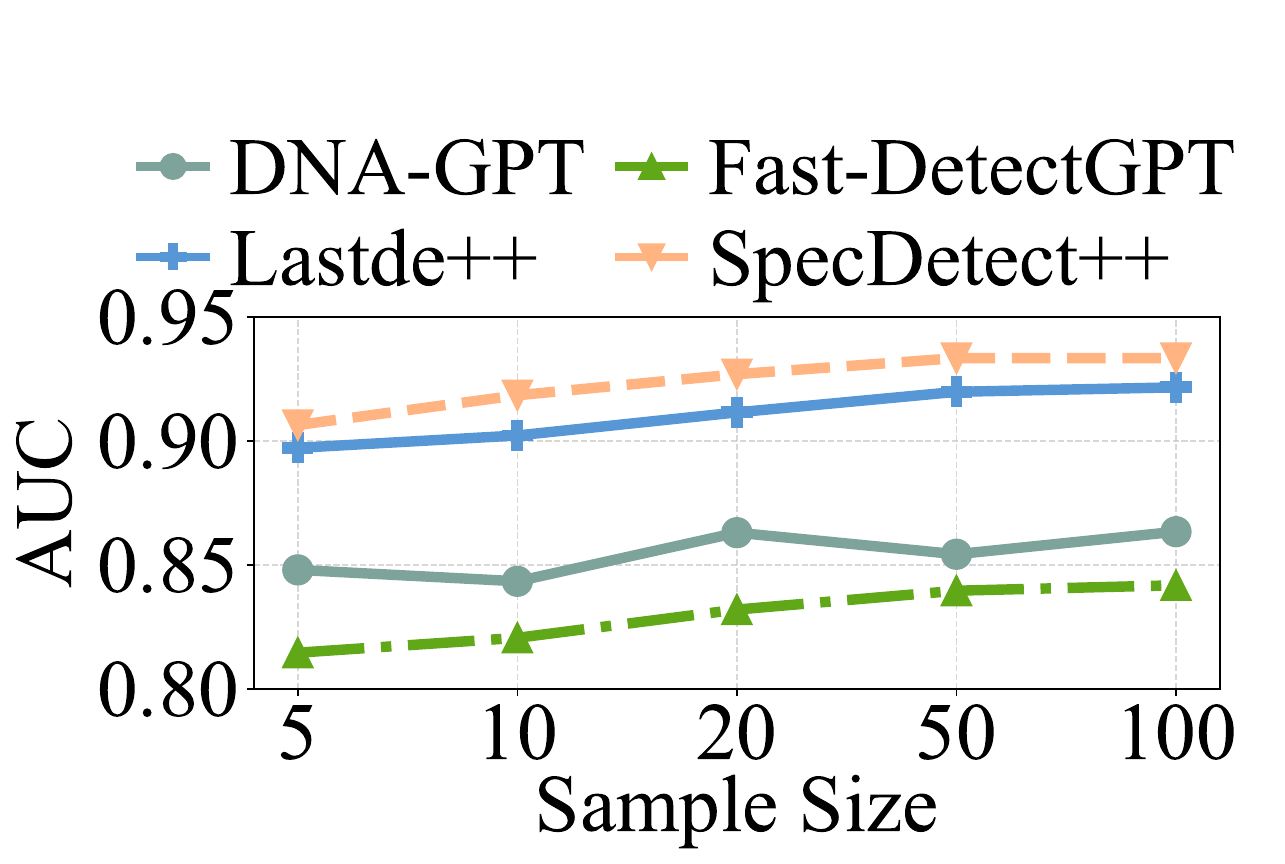}
        \label{fig:sample_size}
    }
    \caption{Analysis of detector performance as a function of (a) text length and (b) number of contrast samples.}
    \label{fig:detector_analysis}
\end{figure}

\paragraph{Impact of Contrast Sample Size.}
For distribution-based methods, we analyze the impact of the number of contrast samples on performance. We conduct experiments using the BLOOM-7.1B source model, with results averaged across three datasets. The results are in  Figure~\ref{fig:sample_size} and detailed results in Appendix \ref{app:sample_size}. From the results, SpecDetect++ not only consistently outperforms all baselines across all sample sizes but also demonstrates remarkable sample efficiency. Notably, SpecDetect++ with only 10 contrast samples already surpasses the performance of Lastde++ with 50 samples, and nearly matches its performance with 100 samples. This superior sample efficiency highlights the strong and stable discriminative power of our underlying spectral energy feature, which allows for robust detection with significantly fewer samples, further enhancing its efficiency.


\paragraph{Generalization to Different Proxy Models.}

\begin{figure}[t]
    \centering
    \includegraphics[scale=0.3]{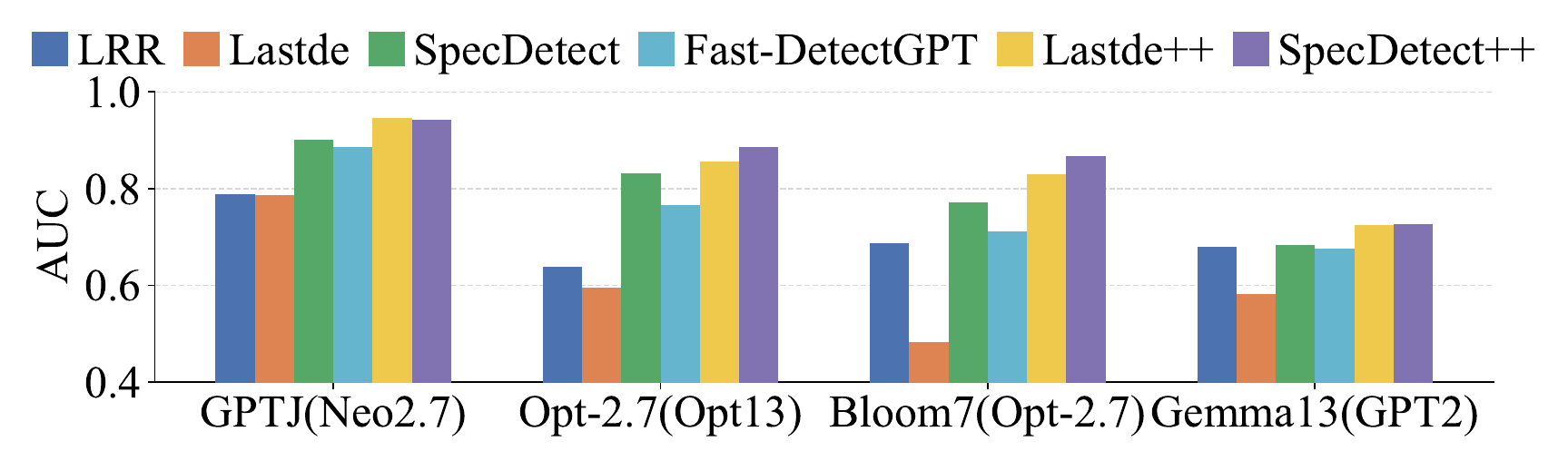}
    \caption{AUC performance with different source and proxy model pairs on the XSum dataset. Each group of bars represents a source-proxy combination.}
    \label{fig:proxy_model_results}
\end{figure}




Proxy model choice critically impacts black-box detection. We investigate this by evaluating detectors on XSum across four different source-proxy pairs, with results in Figure~\ref{fig:proxy_model_results}. While performance varies significantly, confirming the challenge, our methods demonstrate superior generalization. Our SpecDetect consistently outperforms baselines by a large margin, indicating its spectral feature is more robust to proxy mismatch than complex time-series features. Similarly, SpecDetect++ consistently outperforms other distribution-based methods, showcasing its robustness across all tested pairs.

\subsubsection{Performance on Non-English Datasets.}

\begin{figure}[t]
    \centering
    \includegraphics[scale=0.42]{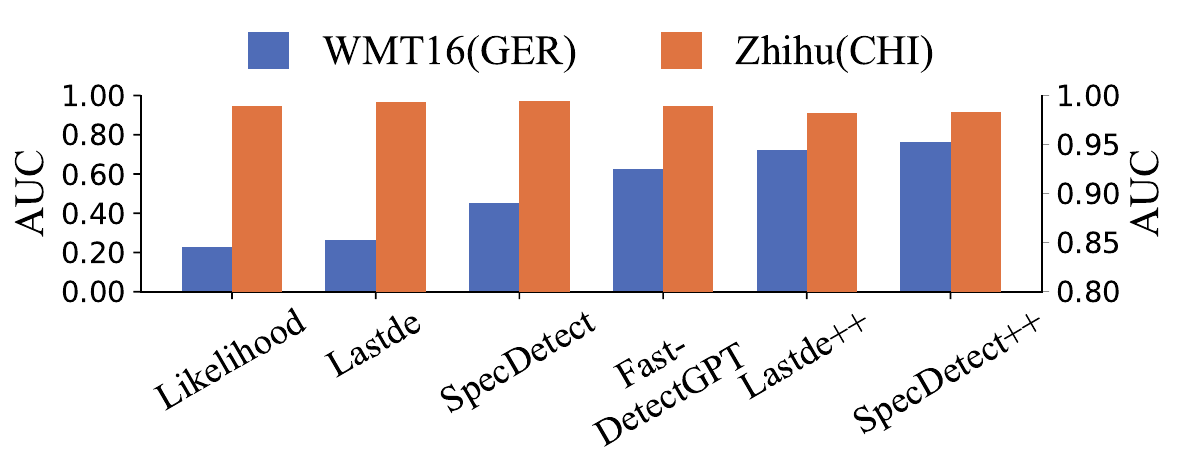}
    \caption{Performance in two non-English scenarios: a German task on WMT16-De (left y-axis; source: mGPT, proxy: GPT-J) and a Chinese task on Zhihu-Economy (right y-axis; source: Qwen-8B, proxy: Yi-1.5-6B).}
    \label{fig:non_english}
\end{figure}


To evaluate generalizability beyond English, we conduct German and Chinese experiments, shown in Figure~\ref{fig:non_english}. Following previous work \cite{xu2024training}, we test German on WMT16 (source: mGPT, proxy: GPT-J) and Chinese on an economics dataset (source: Qwen-7B, proxy: Yi-1.5-6B). The results show our framework's strong cross-lingual performance. Our base method, SpecDetect, achieves top-tier AUC. The enhanced SpecDetect++ is also highly competitive, outperforming the SOTA Lastde++ in the German scenario, highlighting our approach's effectiveness in non-English settings.

\section{Conclusion}

\label{sec:conclusion}

We introduce SpecDetect, a novel, training-free LLM text detector using a frequency-domain perspective. By treating the token probability sequence as a signal, it uses a single, hyperparameter-free feature, DFT Total Energy, to robustly distinguish human from machine text. This method is simple and efficient. Experiments confirm SpecDetect set a new SOTA in effectiveness-efficiency trade-off, achieving superior or comparable performance at half the computational cost with strong robustness. This work advances LLM text detection via classical signal processing, and future research can explore more spectral features for fine-grained tasks.

\section*{Acknowledgments}
This work was supported in whole or in part, by National Natural Science Foundation of China (U24B6012, U2333201, 62372429), the Innovation Funding of ICT, CAS under Grant No. E461040, Pilot for Major Scientific Research Facility of Jiangsu Province of China (No.BM2021800).

\bibliography{aaai2026}

\clearpage

\clearpage

\appendix

\section{Experiment Setup Details}
\subsection{Baseline Descriptions}
\label{app:baselines}
This section details the implementation of the 10 baseline methods used for comparison. These methods are categorized into two groups: sample-based methods and distribution-based methods. For all black-box scenarios, we use \textbf{GPT-J} as the default scoring, sampling, and rewriting model unless otherwise specified.

\paragraph{Sample-based Methods.}
This first group of methods computes a detection score by aggregating probability information for each token in the given text.

\begin{itemize}
    \item \textbf{Log-Likelihood} \cite{solaiman2019release}: The metric is the average log probability of all tokens in the candidate text.
    \item \textbf{Log-Rank} \cite{solaiman2019release}: This method uses the average log rank of each token, where ranks are determined by sorting the vocabulary in descending order of probability.
    \item \textbf{Entropy} \cite{gehrmann2019gltr,ippolito2019automatic}: The metric is the average token entropy, calculated from the probability distribution over the entire vocabulary at each position.
    \item \textbf{DetectLRR} \cite{su2023detectllm}: This method calculates the ratio of the log-likelihood to the log-rank for each token and uses the average ratio as the final detection score.
    \item \textbf{Lastde} \cite{xu2024training}: This method treats the token probability sequence as a time series and uses Multiscale Diversity Entropy (MDE) to measure its complexity and fluctuations. Adopting the default configuration, we set the hyperparameters as follows:  $s=3$,  $\epsilon=10\times n$, and $\tau^{'}=5$, with $n$ representing the number of tokens in the text.
\end{itemize}

\paragraph{Distribution-based Methods.}
This second group of methods generates a set of contrastive samples through perturbation or sampling. The final detection score is derived from the statistical discrepancy between the original text and these generated samples.

\begin{itemize}
    \item \textbf{DetectGPT} \cite{mitchell2023detectgpt}: This method uses the discrepancy in likelihood after perturbation as its metric. We followed the original paper's default settings, employing \textbf{T5-3B} as the perturbation model and generating \textbf{100} perturbations.
    
    \item \textbf{DetectNPR} \cite{su2023detectllm}: This baseline uses normalized log-rank perturbation as its metric. All other settings are kept consistent with DetectGPT.
    
    \item \textbf{DNA-GPT} \cite{yang2023dna}: Contrast samples are generated using a ``cut-off and rewrite" technique. The metric is the log-likelihood. We adopt the default settings of a truncation rate of 0.5 and \textbf{10} rewrites.
    
    \item \textbf{Fast-DetectGPT} \cite{bao2023fast}: This method uses sampling discrepancy as its metric. To ensure a fair comparison, we use \textbf{GPT-J} as both the sampling and scoring model, deviating from the original paper's use of GPT-J for sampling and Neo-2.7 for scoring. The number of samples is maintained at \textbf{10,000}.

    \item \textbf{Lastde++} \cite{xu2024training}: This method extends Lastde by incorporating a sampling discrepancy mechanism, using the z-score of the original text's Lastde score relative to the distribution of scores from sampled texts. 
     Following the default configuration, we generate \textbf{100} contrastive samples and set the hyperparameters as: $s=4$, $\epsilon=8\times n$, and $\tau^{'}=15$, where $n$ denotes the number of tokens in the text. 
    
\end{itemize}

\subsection{Dataset Details}
\label{app:dataset}
Following the configurations in prior studies \cite{bao2023fast,yang2023dna,xu2024training}, we evaluate our approach using four English datasets: (i) \textbf{XSum} \cite{narayan2018don}, which contains BBC News articles for summarization tasks; (ii) \textbf{SQuAD} \cite{rajpurkar2016squad,rajpurkar2018know}, comprising Wikipedia-based question-answering contexts; (iii) \textbf{WritingPrompts} \cite{fan2018hierarchical}, featuring creative story generation prompts; and (iv) \textbf{Reddit ELI5} \cite{fan2019eli5}, an explanatory question-answering dataset covering biology, physics, chemistry, economics, law, and technology. To assess cross-lingual robustness, following~\cite{xu2024training}, we include two non-English datasets: (i) \textbf{WMT16-De} \cite{bojar2016findings}, a German dataset originally used for translation tasks; and \textbf{Zhihu-Economy}\cite{xu2024training}, a Chinese question-answering dataset focusing on economics. Following previous work, we use Xsum, SQuAD, and WritingPrompt for white-box detection, while Xsum, Reddit ELI5, and WritingPrompts for black-box detection.  Each dataset includes 150 human-written examples. For LLM-generated text, we use the first 30 tokens of each human-written example as a prompt, with continuations generated by source models. As a result, each finalized dataset comprises 150 negative samples (human-written) and 150 positive samples (LLM-generated), with each pair featuring the same text length and matching initial segments.

\subsection{Implementation Details}
\label{app:imple}
Our experiments are conducted using a single 80GB NVIDIA H800 GPU. The models employed in our work are listed in Table \ref{tab:model_specs}. While the majority of these models are consistent with those used in prior research \cite{xu2024training}, we have replaced some with the latest alternatives to verify the effectiveness of our methods on newly released models.

\begin{table*}[t]
\centering
\caption{Specification of models employed in experiments}
\label{tab:model_specs}
\scalebox{0.80}{
\begin{tabular}{lcccc}
\toprule
\textbf{Model Name} & \textbf{Source/Repository} & \textbf{Parameter Scale} & \textbf{Type} & \textbf{Release Year} \\
\midrule
GPT-2 \cite{radford2019language} & openai-community/gpt2-xl & 1.5B & Open-source & 2019 \\
GPT-Neo-2.7 \cite{gpt-neo} & EleutherAI/gpt-neo-2.7B & 2.7B & Open-source & 2021 \\
OPT-2.7 \cite{zhang2022opt} & facebook/opt-2.7b & 2.7B & Open-source & 2022 \\
LLaMA-13 \cite{touvron2023llama1} & huggyllama/llama-13b & 13B & Open-source & 2023 \\
LLaMA2-13 \cite{touvron2023llama2} & TheBloke/Llama-2-13B-fp16 & 13B & Open-source & 2023 \\
LLaMA3-8 \cite{llama3modelcard} & meta-llama/Meta-Llama-3-8B & 8B & Open-source & 2024 \\
OPT-13 \cite{zhang2022opt} & facebook/opt-13b & 13B & Open-source & 2022 \\
BLOOM-7.1 \cite{workshop2022bloom} & bigscience/bloom-7b1 & 7.1B & Open-source & 2022 \\
Phi-4 \cite{abdin2024phi} & microsoft/phi-4 & 4B & Open-source & 2024 \\
Qwen3-8B \cite{qwen3technicalreport} & Qwen/Qwen3-8B & 8B & Open-source & 2025 \\
Gemma3-1B \cite{gemma_2025} & google/gemma-3-1b-it & 1B & Open-source & 2025 \\
GPT-J-6B \cite{gpt-j} & EleutherAI/gpt-j-6b & 6B & Open-source & 2021 \\
Yi1.5-6B \cite{young2024yi} & 01-ai/Yi-1.5-6B & 6B & Open-source & 2024 \\
Claude-3-haiku \cite{claude3_2024} & Anthropic (API) & NA & Closed-source & 2024 \\
GPT-4-Turbo \cite{achiam2023gpt} & OpenAI (API) & NA & Closed-source & 2024 \\
\bottomrule
\end{tabular}
}
\end{table*}

\section{Complete Experiment Results}
\subsection{Visualization of STFT Spectrum}
\label{app:visual_stft}
\begin{figure}[t]
    \centering

    \begin{minipage}[t]{0.48\linewidth}
        \centering
        \includegraphics[width=\linewidth]{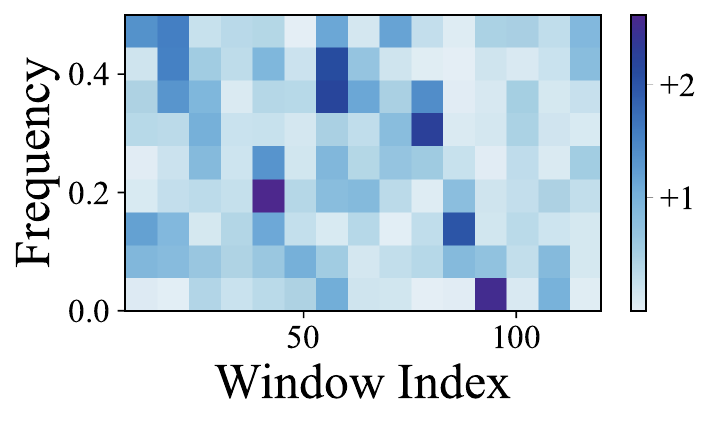}
        {\small (a) Human Text}
    \end{minipage}
    \hfill
    \begin{minipage}[t]{0.48\linewidth}
        \centering
        \includegraphics[width=\linewidth]{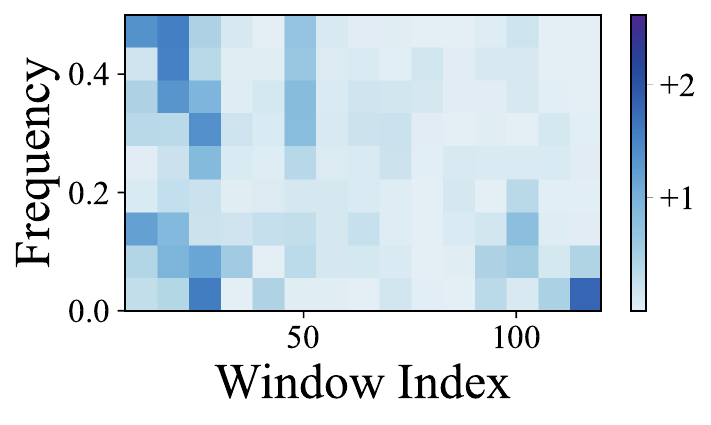}
        {\small (b) LLM-Generated Text}
    \end{minipage}

    \caption{STFT spectrograms (heatmaps) of representative human (a) and LLM-generated (b) text from the WritingPrompts dataset. Differences in spectral patterns reflect variation in local frequency structure.}
    \label{fig:stft_heatmap}
\end{figure}

For a qualitative comparison, we present the Short-Time Fourier Transform (STFT) heatmap for a representative human-LLM sample pair in Figure~\ref{fig:stft_heatmap}. The visualization reveals that human text exhibits more irregular, high-intensity ``hotspots" across the time-frequency space, indicating abrupt changes in signal dynamics. In contrast, the LLM-generated text maintains more uniform and low-variability patterns throughout.

\subsection{Black-Box Detection Performance}
\label{app:black_box}
We present the detailed results of the black-box experiments in Table \ref{tab:black_xsum}, Table \ref{tab:black_writing} and Table \ref{tab:black_reddit}. In line with the analysis in the main text, these results further validate the stable performance of our method under various experimental settings, reinforcing that our proposed framework maintains strong adaptability in complex black-box environments.

\begin{table*}
\centering
\resizebox{\linewidth}{!}{
\begin{tabular}{lcccccccccccc}
\toprule
\textbf{Method} & \textbf{GPT-2} & \textbf{Neo-2.7} & \textbf{OPT-2.7} & \textbf{LLaMA-13} & \textbf{LLaMA2-13} & \textbf{LLaMA3-8} & \textbf{OPT-13} & \textbf{BLOOM-7.1} & \textbf{Phi-4} & \textbf{Qwen3-8} & \textbf{Claude-3} & \textbf{GPT4-Turbo} \\
\midrule
\multicolumn{13}{c}{\textit{sample-based}} \\ \hline
Likelihood & 0.5401 & 0.4846 & 0.6321 & 0.5424 & 0.5435 & 0.9891 & 0.6884 & 0.3949 & 0.5319 & \underline{0.9683} & 0.9684 & 0.6047 \\
LogRank & 0.5833 & 0.5353 & 0.6687 & 0.6009 & 0.5938 & \underline{0.9912} & 0.7226 & 0.4720 & 0.5643 & \textbf{0.9760} & \underline{0.9706} & 0.6148 \\
Entropy & 0.6547 & 0.6986 & 0.5684 & 0.5376 & 0.5156 & 0.2908 & 0.5323 & 0.7332 & 0.5778 & 0.4772 & 0.3888 & 0.6125 \\
DetectLRR & 0.6727 & 0.6668 & 0.7161 & \underline{0.7172} & \textbf{0.6811} & 0.9623 & 0.7482 & 0.6608 & \underline{0.6204} & 0.9612 & 0.9575 & 0.6164 \\
Lastde & \textbf{0.8174} & \textbf{0.8336} & \underline{0.8354} & \textbf{0.7304} & \underline{0.6790} & \textbf{0.9933} & \underline{0.8633} & \textbf{0.7814} & 0.5558 & 0.9355 & \textbf{0.9747} & \underline{0.6907} \\     
\textbf{SpecDetect} & \underline{0.8168} & \underline{0.8324} & \textbf{0.8649} & 0.6972 & 0.6716 & 0.9730 & \textbf{0.8681} & \underline{0.7367} & \textbf{0.6455} & 0.9575 & 0.9652 & \textbf{0.7120} \\

\hline
\multicolumn{13}{c}{\textit{distribution-based}} \\ \hline
DetectGPT & 0.5925 & 0.6950 & 0.6200 & 0.5100 & 0.5600 & 0.7700 & 0.7300 & 0.3325 & 0.5900 & 0.7400 & 0.8625 & 0.4675 \\
DetectNPR & 0.5900 & 0.6425 & 0.6450 & 0.5250 & 0.5750 & 0.9775 & 0.6425 & 0.3775 & 0.6250 & 0.8650 & 0.8750 & 0.5200 \\
DNA-GPT & 0.4965 & 0.4379 & 0.5789 & 0.5211 & 0.5444 & 0.9891 & 0.6040 & 0.4704 & 0.5575 & 0.9500 & 0.9425 & 0.5925 \\
Fast-DetectGPT & 0.8080 & 0.7721 & 0.8208 & 0.6480 & 0.6283 & \textbf{0.9919} & 0.8426 & 0.6950 & 0.6514 & \textbf{0.9845} & \underline{0.9993} & 0.8077 \\
Lastde++ & \underline{0.8867} & \underline{0.8942} & \underline{0.9077} & \underline{0.7349} & \underline{0.7280} & \underline{0.9880} & \textbf{0.9177} & \underline{0.8196} & \underline{0.7363} & 0.9670 & \textbf{0.9995} & \textbf{0.8272} \\
\textbf{SpecDetect++} & \textbf{0.8988} & \textbf{0.9119} & \textbf{0.9149} & \textbf{0.7592} & \textbf{0.7563} & 0.9645 & \underline{0.9105} & \textbf{0.8520} & \textbf{0.7367} & \underline{0.9775} & 0.9849 & \underline{0.8210} \\
\bottomrule
\end{tabular}
} 
\caption{Detection AUC under black box scenarios on the XSum dataset.}
\label{tab:black_xsum}
\end{table*}

\begin{table*}
\centering
\resizebox{\linewidth}{!}{
\begin{tabular}{lcccccccccccc}
\toprule
\textbf{Method} & \textbf{GPT-2} & \textbf{Neo-2.7} & \textbf{OPT-2.7} & \textbf{LLaMA-13} & \textbf{LLaMA2-13} & \textbf{LLaMA3-8} & \textbf{OPT-13} & \textbf{BLOOM-7.1} & \textbf{Phi-4} & \textbf{Qwen3-8} & \textbf{Claude-3} & \textbf{GPT4-Turbo} \\
\midrule
\multicolumn{13}{c}{\textit{sample-based}} \\ \hline
Likelihood & 0.7978 & 0.8236 & 0.7707 & 0.7644 & 0.7897 & 0.9298 & 0.7743 & 0.7676 & 0.8132 & 0.9922 & 0.9838 & \textbf{0.8148} \\
LogRank & 0.8325 & 0.8570 & 0.8197 & 0.7976 & 0.8176 & 0.9455 & 0.8140 & 0.8144 & 0.8184 & \textbf{0.9969} & 0.9871 & \underline{0.7895} \\
Entropy & 0.5325 & 0.4895 & 0.4868 & 0.4504 & 0.3968 & 0.2372 & 0.4488 & 0.5243 & 0.3576 & 0.2341 & 0.1769 & 0.3556 \\
DetectLRR & 0.8842 & 0.9044 & 0.9062 & 0.8533 & 0.8578 & 0.9489 & 0.8796 & 0.9007 & 0.7950 & 0.9884 & 0.9796 & 0.6672 \\
Lastde & \underline{0.9615} & \underline{0.9676} & \underline{0.9634} & \underline{0.8778} & \underline{0.8785} & \textbf{0.9631} & \textbf{0.9442} & \underline{0.9593} & \underline{0.8233} & 0.9951 & \textbf{0.9940} & 0.6882 \\
\textbf{SpecDetect} & \textbf{0.9655} & \textbf{0.9756} & \textbf{0.9640} & \textbf{0.8892} & \textbf{0.8843} & \underline{0.9622} & \underline{0.9420} & \textbf{0.9610} & \textbf{0.8489} & \underline{0.9965} & \underline{0.9912} & 0.7849 \\

\hline
\multicolumn{13}{c}{\textit{distribution-based}} \\ \hline
DetectGPT & 0.7575 & 0.7950 & 0.8275 & 0.7550 & 0.8650 & 0.7325 & 0.8150 & 0.6875 & 0.8650 & 0.7725 & 0.9925 & \textbf{0.9050} \\
DetectNPR & 0.8400 & 0.8775 & 0.8700 & 0.8025 & 0.8625 & 0.8875 & 0.8500 & 0.7425 & 0.8725 & 0.9500 & \textbf{1.0000} & 0.8750 \\
DNA-GPT & 0.7366 & 0.7612 & 0.7298 & 0.6774 & 0.7361 & 0.8852 & 0.7365 & 0.7335 & 0.7900 & 0.9675 & \textbf{1.0000} & 0.7275 \\
Fast-DetectGPT & 0.9542 & 0.9606 & 0.9202 & 0.8825 & 0.8641 & 0.9299 & 0.8963 & 0.9276 & 0.8672 & \textbf{0.9980} & \underline{0.9999} & \underline{0.8990} \\
Lastde++ & \underline{0.9816} & \underline{0.9878} & \underline{0.9712} & \underline{0.9421} & \underline{0.9255} & \textbf{0.9705} & \underline{0.9543} & \underline{0.9751} & \textbf{0.8985} & 0.9899 & 0.9996 & 0.8858 \\
\textbf{SpecDetect++} & \textbf{0.9822} & \textbf{0.9889} & \textbf{0.9782} & \textbf{0.9469} & \textbf{0.9267} & \underline{0.9634} & \textbf{0.9601} & \textbf{0.9797} & \underline{0.8765} & \underline{0.9970} & 0.9960 & 0.8940 \\
\bottomrule
\end{tabular}
} 
\caption{Detection AUC under black box scenarios on the WritingPrompts dataset.}
\label{tab:black_writing}
\end{table*}

\begin{table*}
\centering
\resizebox{\linewidth}{!}{
\begin{tabular}{lcccccccccccc}
\toprule
\textbf{Method} & \textbf{GPT-2} & \textbf{Neo-2.7} & \textbf{OPT-2.7} & \textbf{LLaMA-13} & \textbf{LLaMA2-13} & \textbf{LLaMA3-8} & \textbf{OPT-13} & \textbf{BLOOM-7.1} & \textbf{Phi-4} & \textbf{Qwen3-8} & \textbf{Claude-3} & \textbf{GPT4-Turbo} \\
\midrule
\multicolumn{13}{c}{\textit{sample-based}} \\ \hline
Likelihood & 0.6555 & 0.7047 & 0.6192 & 0.6657 & 0.7252 & \textbf{1.0000} & 0.6014 & 0.6916 & 0.8104 & \textbf{0.9908} & \underline{0.9992} & \underline{0.9714} \\
LogRank & 0.7069 & 0.7426 & 0.6800 & 0.7105 & 0.7679 & \textbf{1.0000} & 0.6524 & 0.7382 & 0.8159 & \underline{0.9883} & \textbf{0.9996} & \textbf{0.9718} \\
Entropy & 0.6496 & 0.5716 & 0.5810 & 0.4858 & 0.4427 & 0.0805 & 0.6114 & 0.5678 & 0.3380 & 0.1406 & 0.0656 & 0.0848 \\
DetectLRR & 0.8179 & 0.8050 & 0.8131 & 0.7845 & 0.8257 & 0.9696 & 0.7760 & 0.8251 & 0.7969 & 0.9635 & 0.9959 & 0.9315 \\
Lastde & \underline{0.9243} & \underline{0.9107} & \underline{0.8978} & \underline{0.8122} & \underline{0.8483} & \underline{0.9990} & \underline{0.8894} & \underline{0.9296} & \underline{0.8269} & 0.9477 & \underline{0.9992} & 0.9039 \\
\textbf{SpecDetect} & \textbf{0.9412} & \textbf{0.9432} & \textbf{0.9064} & \textbf{0.8288} & \textbf{0.8710} & 0.9931 & \textbf{0.8915} & \textbf{0.9366} & \textbf{0.8374} & 0.9808 & 0.9934 & 0.9193 \\

\hline
\multicolumn{13}{c}{\textit{distribution-based}} \\ \hline
DetectGPT & 0.7300 & 0.8150 & 0.5850 & 0.6800 & 0.6525 & 0.7625 & 0.6575 & 0.7100 & 0.8650 & 0.9700 & 0.9375 & 0.8850 \\
DetectNPR & 0.7100 & 0.7850 & 0.5300 & 0.7150 & 0.7125 & 0.9700 & 0.6975 & 0.7900 & 0.7750 & \textbf{1.0000} & 0.9875 & 0.9325 \\
DNA-GPT & 0.6518 & 0.6894 & 0.5999 & 0.6102 & 0.7144 & \textbf{1.0000} & 0.6925 & 0.6725 & 0.8275 & 0.9500 & 0.9350 & 0.7400 \\
Fast-DetectGPT & 0.9279 & 0.9296 & 0.8543 & 0.7970 & 0.8360 & \underline{0.9953} & 0.8459 & 0.9148 & \underline{0.8791} & 0.9807 & \textbf{0.9996} & \textbf{0.9388} \\
Lastde++ & \underline{0.9705} & \textbf{0.9808} & \underline{0.9395} & \underline{0.8781} & \textbf{0.9133} & 0.9941 & \underline{0.9330} & \textbf{0.9699} & \textbf{0.9253} & 0.9598  & \underline{0.9992} & 0.9325 \\
\textbf{SpecDetect++} & \textbf{0.9744} & \underline{0.9754} & \textbf{0.9463} & \textbf{0.8846} & \underline{0.9100} & 0.9785 & \textbf{0.9374} & \underline{0.9682} & 0.8743 & \underline{0.9856} & 0.9878 & \underline{0.9354} \\
\bottomrule
\end{tabular}
} 
\caption{Detection AUC under black box scenarios on the Reddit dataset.}
\label{tab:black_reddit}
\end{table*}

\subsection{White-box Detection Performance}
\label{app:white_box}

Table \ref{tab:whitebox_results} presents the average AUC across three datasets (XSum, SQuAD, and WritingPrompts), with the detailed, dataset-specific results provided in Tables \ref{tab:white_xsum}, \ref{tab:white_squad}, and \ref{tab:white_writing}. While the white-box setting is crucial for a direct comparison of methods, it is less representative of real-world detection scenarios than black-box testing. In this context, our SpecDetect method sets a new state-of-the-art in the sample-based category, achieving the highest average AUC of 0.9682. This performance, which surpasses all other competitors, is particularly strong on the challenging LLaMA and LLaMA2 models, highlighting the robustness of our frequency-domain approach. Although SpecDetect++ places a close second in the distribution-based category, its competitive performance on specific models validates the power of our novel frequency-domain features, which capture statistical patterns in a way distinct from traditional metrics. These results provide a strong foundation for our method's superior performance in the more realistic black-box scenarios discussed in the main body of the paper.

\begin{table*}
\centering
\resizebox{\linewidth}{!}{
\begin{tabular}{lccccccccccccc}
\toprule
\textbf{Method} & \textbf{GPT-2} & \textbf{Neo-2.7} & \textbf{OPT-2.7} & \textbf{LLaMA-13} & \textbf{LLaMA2-13} & \textbf{LLaMA3-8} & \textbf{OPT-13} & \textbf{BLOOM-7.1} & \textbf{Phi-4} & \textbf{Qwen3-8} & \textbf{Gemma3-1}  & \textbf{GPTJ-6B} & \textbf{Avg.}\\
\midrule
\multicolumn{13}{c}{\textit{sample-based}} \\ \hline
Likelihood       & 0.9193 & 0.8940 & 0.8808 & 0.6366 & 0.6536 & 0.9818 & 0.8446 & 0.8800 & 0.8092 & \underline{0.9872} & \textbf{0.9967} & 0.8496 & 0.8611 \\
LogRank  & 0.9437 & 0.9287 & 0.9099 & 0.6887 & 0.7030 & 0.9861 & 0.8776 & 0.9242 & 0.8359 & \textbf{0.9891} & \underline{0.9966} & 0.8870 & 0.8892 \\
Entropy  & 0.5201 & 0.5172 & 0.5046 & 0.6415 & 0.6105 & 0.1755 & 0.5431 & 0.6266 & 0.4656 & 0.2905 & 0.1953 & 0.5430 & 0.4695 \\
DetectLRR  & 0.9632 & 0.9607 & 0.9313 & 0.8147 & 0.8097 & 0.9676 & 0.9095 & 0.9633 & 0.8450 & 0.8937 & 0.8950 & 0.9227 & 0.9064 \\
Lastde  & \underline{0.9826} & \textbf{0.9865} & \underline{0.9815} & \underline{0.8953} & \underline{0.8876} & \textbf{0.9952} & \underline{0.9692} & \textbf{0.9943} & \underline{0.8608} & 0.9821 & \textbf{0.9967} & \underline{0.9711} & \underline{0.9586} \\
\textbf{SpecDetect}  & \textbf{0.9869} & \underline{0.9856} & \textbf{0.9863} & \textbf{0.9048} & \textbf{0.9073} & \underline{0.9885} & \textbf{0.9745} & \underline{0.9900} & \textbf{0.9409} & 0.9844 & 0.9927 & \textbf{0.9760} & \textbf{0.9682} \\ \hline
\multicolumn{13}{c}{\textit{distribution-based}} \\ \hline
DetectGPT  &0.9750	&0.9733	&0.9808	&0.6583	&0.7117	&0.6442	&0.8975	&0.9467	&0.8608	&0.9225	&\textbf{1.0000}	&0.9142 & 0.8737 \\
DetectNPR  &0.9808	&0.9733	&0.9658	&0.7000	&0.7350	&0.9217	&0.9358	&0.9667	&0.8508	&\underline{0.9925}	&\textbf{1.0000}	&0.9175 & 0.9117 \\
DNA-GPT  &0.9090	&0.8712	&0.8638	&0.6210	&0.6494	&0.9487	&0.8317	&0.8787	&0.7142	&\textbf{0.9983}	&\textbf{1.0000}	&0.8825 & 0.8474 \\
Fast-DetectGPT & \underline{0.9957} & 0.9949 & 0.9878 & 0.9342 & 0.9335 & \underline{0.9953} & 0.9822 & 0.9959 & \underline{0.9324} & 0.9811 & 0.9978 & \underline{0.9894} & 0.9767 \\
Lastde++  & \textbf{0.9974} & \textbf{0.9987} & \textbf{0.9946} & \textbf{0.9665} & \underline{0.9672} & \textbf{0.9973} & \textbf{0.9917} & \textbf{0.9993} & \textbf{0.9493} & 0.9810 & \underline{0.9980} & \textbf{0.9948} & \textbf{0.9863} \\
\textbf{SpecDetect++}  & 0.9926 & \underline{0.9958} & \underline{0.9921} & \underline{0.9664} & \textbf{0.9699} & 0.9847 & \underline{0.9845} & \underline{0.9971} & 0.9211 & 0.9709 & 0.9629 & 0.9889 & \underline{0.9772} \\
\bottomrule
\end{tabular}
} 
\caption{Detection AUC under white box scenarios, with values averaged across three datasets: XSum, SQuAD, and WritingPrompts. The last column (``Avg.") denotes the mean AUC across all source models, calculated over the three datasets. Best and second-best results are highlighted in \textbf{bold} and \underline{underline}, respectively.}
\label{tab:whitebox_results}
\end{table*}

\begin{table*}
\centering
\resizebox{\linewidth}{!}{
\begin{tabular}{lcccccccccccc}
\toprule
\textbf{Method} & \textbf{GPT-2} & \textbf{Neo-2.7} & \textbf{OPT-2.7} & \textbf{LLaMA-13} & \textbf{LLaMA2-13} & \textbf{LLaMA3-8} & \textbf{OPT-13} & \textbf{BLOOM-7.1} & \textbf{Phi-4} & \textbf{Qwen3-8} & \textbf{Gemma3-1}  & \textbf{GPTJ-6B} \\
\midrule
\multicolumn{13}{c}{\textit{sample-based}} \\ \hline
Likelihood & 0.8689 & 0.8668 & 0.8146 & 0.6165 & 0.6308 & 0.9957 & 0.7717 & 0.8530 & 0.7458 & \underline{0.9864} & \textbf{0.9934} & 0.8146 \\
LogRank & 0.9000 & 0.9059 & 0.8464 & 0.6706 & 0.6880 & \underline{0.9974} & 0.8105 & 0.9066 & 0.7852 & \textbf{0.9884} & \textbf{0.9934} & 0.8496 \\
Entropy & 0.5619 & 0.5982 & 0.5433 & 0.6629 & 0.6298 & 0.1992 & 0.5737 & 0.6956 & 0.5678 & 0.3569 & 0.2236 & 0.5977 \\
DetectLRR & 0.9249 & 0.9244 & 0.8557 & 0.7925 & 0.7696 & 0.9770 & 0.8442 & 0.9516 & 0.7974 & 0.9383 & 0.8676 & 0.8660 \\
Lastde & \underline{0.9591} & \textbf{0.9747} & \underline{0.9567} & \underline{0.8913} & \underline{0.8592} & \textbf{1.0000} & \underline{0.9402} & \textbf{0.9926} & \underline{0.8136} & 0.9720 & \textbf{0.9934} & \textbf{0.9623} \\
\textbf{SpecDetect} & \textbf{0.9668} & \underline{0.9709} & \textbf{0.9707} & \textbf{0.9277} & \textbf{0.9102} & 0.9912 & \textbf{0.9443} & \underline{0.9795} & \textbf{0.9190} & 0.9769 & \underline{0.9898} & \underline{0.9567} \\
 \hline
\multicolumn{13}{c}{\textit{distribution-based}} \\ \hline
DetectGPT & 0.9300 & 0.9575 & 0.9625 & 0.6925 & 0.7550 & 0.7400 & 0.8525 & 0.9350 & 0.8000 & 0.9675 & \textbf{1.0000} & 0.9200 \\
DetectNPR & 0.9425 & 0.9450 & 0.9125 & 0.6525 & 0.7950 & 0.9900 & 0.8725 & 0.9675 & 0.8750 & \textbf{1.0000} & \textbf{1.0000} & 0.9174 \\
DNA-GPT & 0.8484 & 0.8071 & 0.7710 & 0.5997 & 0.6064 & 0.9789 & 0.7253 & 0.8032 & 0.6825 & \textbf{1.0000} & \textbf{1.0000} & 0.7562 \\
Fast-DetectGPT & \underline{0.9891} & 0.9896 & 0.9698 & 0.9480 & 0.9263 & \textbf{0.9990} & 0.9548 & 0.9921 & \underline{0.9340} & \underline{0.9678} & 0.9934 & \underline{0.9812} \\
Lastde++ & \textbf{0.9932} & \textbf{0.9972} & \textbf{0.9849} & \textbf{0.9754} & \textbf{0.9607} & \underline{0.9971} & \textbf{0.9776} & \textbf{0.9984} & \textbf{0.9570} & 0.9671 & \underline{0.9970} & \textbf{0.9900} \\
\textbf{SpecDetect++} & 0.9832 & \underline{0.9932} & \underline{0.9804} & \underline{0.9707} & \underline{0.9572} & 0.9889 & \underline{0.9603} & \underline{0.9942} & 0.9287 & 0.9596 & 0.9772 & 0.9799 \\
\bottomrule
\end{tabular}
} 
\caption{Detection AUC under white box scenarios on the XSum dataset.}
\label{tab:white_xsum}
\end{table*}

\begin{table*}
\centering
\resizebox{\linewidth}{!}{
\begin{tabular}{lcccccccccccc}
\toprule
\textbf{Method} & \textbf{GPT-2} & \textbf{Neo-2.7} & \textbf{OPT-2.7} & \textbf{LLaMA-13} & \textbf{LLaMA2-13} & \textbf{LLaMA3-8} & \textbf{OPT-13} & \textbf{BLOOM-7.1} & \textbf{Phi-4} & \textbf{Qwen3-8} & \textbf{Gemma3-1}  & \textbf{GPTJ-6B} \\
\midrule
\multicolumn{13}{c}{\textit{sample-based}} \\ \hline
Likelihood & 0.9205 & 0.8677 & 0.8884 & 0.4656 & 0.4835 & 0.9861 & 0.8404 & 0.8508 & 0.7363 & 0.9752 & \textbf{1.0000} & 0.8049 \\
LogRank & 0.9514 & 0.9148 & 0.9239 & 0.5185 & 0.5357 & \underline{0.9902} & 0.8793 & 0.9052 & 0.7644 & \textbf{0.9790} & \underline{0.9997} & 0.8590 \\
Entropy & 0.5619 & 0.5982 & 0.5433 & 0.6629 & 0.6298 & 0.1992 & 0.5737 & 0.6956 & 0.5678 & 0.3569 & 0.2236 & 0.5977 \\
DetectLRR & 0.9797 & 0.9692 & 0.9583 & 0.7164 & 0.7348 & 0.9780 & 0.9208 & 0.9527 & 0.7924 & 0.7732 & 0.8865 & 0.9276 \\
Lastde & \underline{0.9962} & \underline{0.9887} & \underline{0.9897} & \textbf{0.8168} & \textbf{0.8336} & \textbf{0.9963} & \underline{0.9780} & \underline{0.9928} & \underline{0.8029} & 0.9742 & \textbf{1.0000} & \underline{0.9653} \\
\textbf{SpecDetect} & \textbf{0.9980} & \textbf{0.9909} & \textbf{0.9945} & \underline{0.8041} & \underline{0.8296} & 0.9866 & \textbf{0.9850} & \textbf{0.9951} & \textbf{0.9253} & \underline{0.9763} & 0.9950 & \textbf{0.9780} \\
 \hline
\multicolumn{13}{c}{\textit{distribution-based}} \\ \hline
DetectGPT & \textbf{1.0000} & 0.9725 & 0.9800 & 0.4400 & 0.5025 & 0.5650 & 0.8450 & 0.9200 & 0.7875 & 0.9050 & \textbf{1.0000} & 0.8400 \\
DetectNPR & \textbf{1.0000} & 0.9750 & 0.9850 & 0.4925 & 0.4775 & 0.8750 & 0.9400 & 0.9400 & 0.6875 & \underline{0.9775} & \textbf{1.0000} & 0.8575 \\
DNA-GPT & 0.9342 & 0.8935 & 0.9054 & 0.5025 & 0.5411 & 0.9746 & 0.8586 & 0.8954 & 0.6425 & \textbf{1.0000} & \textbf{1.0000} & 0.8209 \\
Fast-DetectGPT & 0.9991 & 0.9972 & 0.9969 & 0.8714 & 0.8938 & \textbf{0.9995} & 0.9966 & 0.9975 & \underline{0.9312} & 0.9756 & \underline{0.9999} & \underline{0.9951} \\
Lastde++ & \underline{0.9995} & \textbf{0.9993} & \textbf{0.9995} & \underline{0.9283} & \underline{0.9466} & \underline{0.9986} & \textbf{0.9995} & \textbf{0.9996} & \textbf{0.9396} & 0.9760 & 0.9990 & \textbf{0.9965} \\   
\textbf{SpecDetect++} & 0.9973 & \underline{0.9975} & \underline{0.9987} & \textbf{0.9365} & \textbf{0.9602} & 0.9816 & \underline{0.9973} & \underline{0.9986} & 0.9279 & 0.9696 & 0.9740 & 0.9932 \\

\bottomrule
\end{tabular}
} 
\caption{Detection AUC under white box scenarios on the SQuAD dataset.}
\label{tab:white_squad}
\end{table*}

\begin{table*}
\centering
\resizebox{\linewidth}{!}{
\begin{tabular}{lcccccccccccc}
\toprule
\textbf{Method} & \textbf{GPT-2} & \textbf{Neo-2.7} & \textbf{OPT-2.7} & \textbf{LLaMA-13} & \textbf{LLaMA2-13} & \textbf{LLaMA3-8} & \textbf{OPT-13} & \textbf{BLOOM-7.1} & \textbf{Phi-4} & \textbf{Qwen3-8} & \textbf{Gemma3-1}  & \textbf{GPTJ-6B} \\
\midrule
\multicolumn{13}{c}{\textit{sample-based}} \\ \hline
Likelihood & 0.9686 & 0.9476 & 0.9393 & 0.8276 & 0.8464 & 0.9636 & 0.9216 & 0.9362 & 0.9454 & \textbf{1.0000} & \textbf{1.0000} & 0.9294 \\
LogRank & 0.9796 & 0.9655 & 0.9594 & 0.8770 & 0.8854 & 0.9708 & 0.9430 & 0.9608 & 0.9580 & \textbf{1.0000} & \textbf{1.0000} & 0.9525 \\
Entropy & 0.3997 & 0.3745 & 0.4428 & 0.5572 & 0.4872 & 0.1795 & 0.4454 & 0.5108 & 0.3126 & 0.0880 & 0.0668 & 0.4199 \\
DetectLRR & 0.9850 & 0.9885 & 0.9799 & 0.9352 & 0.9248 & 0.9479 & 0.9636 & 0.9856 & 0.9451 & \underline{0.9695} & 0.9582 & 0.9746 \\
Lastde & \underline{0.9924} & \textbf{0.9960} & \textbf{0.9980} & \underline{0.9779} & \underline{0.9699} & \textbf{0.9892} & \underline{0.9893} & \textbf{0.9975} & \underline{0.9659} & \textbf{1.0000} & \textbf{1.0000} & \underline{0.9856} \\
\textbf{SpecDetect} & \textbf{0.9960} & \underline{0.9951} & \underline{0.9938} & \textbf{0.9826} & \textbf{0.9821} & \underline{0.9876} & \textbf{0.9943} & \underline{0.9953} & \textbf{0.9783} & \textbf{1.0000} & \underline{0.9962} & \textbf{0.9933} \\
 \hline
\multicolumn{13}{c}{\textit{distribution-based}} \\ \hline
DetectGPT & 0.9950 & 0.9900 & \textbf{1.0000} & 0.8425 & 0.8775 & 0.6275 & 0.9950 & 0.9850 & \textbf{0.9950} & 0.8950 & \textbf{1.0000} & 0.9825 \\
DetectNPR & \textbf{1.0000} & \textbf{1.0000} & \textbf{1.0000} & 0.9550 & 0.9325 & 0.9000 & 0.9950 & 0.9925 & \underline{0.9900} & \textbf{1.0000} & \textbf{1.0000} & 0.9775 \\
DNA-GPT & 0.9445 & 0.9130 & 0.9150 & 0.7608 & 0.8006 & 0.8926 & 0.9112 & 0.9375 & 0.8175 & \underline{0.9950} & \textbf{1.0000} & 0.8825 \\
Fast-DetectGPT & 0.9990 & 0.9978 & 0.9968 & 0.9833 & 0.9804 & \underline{0.9874} & 0.9953 & 0.9980 & 0.9320 & \textbf{1.0000} & \textbf{1.0000} & 0.9920 \\
Lastde++ & \underline{0.9996} & \underline{0.9996} & \underline{0.9995} & \textbf{0.9958} & \textbf{0.9943} & \textbf{0.9962} & \textbf{0.9980} & \textbf{0.9998} & 0.9551 & \textbf{1.0000} & \underline{0.9980} & \textbf{0.9979} \\
\textbf{SpecDetect++} & 0.9973 & 0.9968 & 0.9973 & \underline{0.9919} & \underline{0.9924} & 0.9836 & \underline{0.9959} & \underline{0.9986} & 0.9105 & 0.9836 & 0.9376 & \underline{0.9936} \\

\bottomrule
\end{tabular}
} 
\caption{Detection AUC under white box scenarios on the WritingPrompts dataset.}
\label{tab:white_writing}
\end{table*}

\subsection{Robustness to Paraphrasing Attacks}
We provide a detailed, per-dataset breakdown of our method's robustness against adversarial paraphrasing attacks in Table~\ref{app_tab:para_attack}. The results confirm the conclusions from our main analysis, demonstrating that both SpecDetect and SpecDetect++ maintain consistently strong performance and outperform the baselines across all three individual datasets, highlighting the inherent resilience of our DFT total energy feature.

\label{app:para}
\begin{table*}
\centering
\resizebox{\linewidth}{!}{
\begin{tabular}{l|ccc|ccc|ccc}
\toprule 
\textbf{Datasets} & \multicolumn{3}{c|}{\textbf{XSum}} & \multicolumn{3}{c|}{\textbf{WritingPrompts}}  & \multicolumn{3}{c}{\textbf{Reddit}} \\
\textbf{Source Models} & \textbf{Neo-2.7} & \textbf{LLaMA-13}  & \textbf{OPT-13} & \textbf{Neo-2.7} & \textbf{LLaMA-13}  & \textbf{OPT-13} & \textbf{Neo-2.7} & \textbf{LLaMA-13}  & \textbf{OPT-13}\\
\midrule
\hline
DetectLRR & 0.6537 & 0.6786 & 0.7536 & 0.8892 & 0.8489 & 0.8913 & 0.7812 & 0.7684 & 0.7675 \\
Lastde & 0.7677 & 0.6739 & 0.8161 & 0.9413 & 0.8318 & 0.9060 & 0.8768 & 0.7502 & 0.8763 \\
\textbf{SpecDetect} & \underline{0.7992} & 0.6608 & 0.8125 & \underline{0.9517} & 0.8632 & 0.9210 & \underline{0.9315} & 0.7754 & 0.9000 \\\hline
Fast-DetectGPT & 0.7651 & 0.6353 & 0.8180 & \underline{0.9515} & 0.8696 & 0.9116 & 0.9221 & 0.7749 & 0.8781 \\
Lastde++ & \textbf{0.8588} & \underline{0.6865} & \textbf{0.8715} & \textbf{0.9805} & \underline{0.9162} & \textbf{0.9476} & \textbf{0.9730} & \textbf{0.8440} & \underline{0.9306} \\
\textbf{SpecDetect++} & 0.8720 & \textbf{0.7010} & \underline{0.8550} & \textbf{0.9723} & \textbf{0.9239} & \underline{0.9484} & 0.9670 & \underline{0.8538} & \textbf{0.9462} \\
\bottomrule
\end{tabular}
}
\caption{Detection AUC under adversarial paraphrasing attacks in a black-box scenario across three datasets. Best and second-best results are highlighted in \textbf{bold} and \underline{underline}, respectively.}
\label{app_tab:para_attack}
\end{table*}

\subsection{Impact of Text Length}
\label{app:text_length}
We present the detailed results across three datasets for varying text lengths in Table~\ref{app_tab:text_length}. Consistent with the analysis in the main body, these results demonstrate the robust effectiveness of our method across all tested lengths, confirming that our spectral feature is a reliable indicator even for shorter texts.

\begin{table*}
\centering
\resizebox{\linewidth}{!}{
\begin{tabular}{l|cccc|cccc|cccc}
\toprule
\textbf{Datasets} & \multicolumn{4}{c|}{\textbf{XSum}} & \multicolumn{4}{c|}{\textbf{WritingPrompts}} & \multicolumn{4}{c}{\textbf{Reddit}} \\
\textbf{Input Length} & \textbf{60} & \textbf{90} & \textbf{120} & \textbf{150} & \textbf{60} & \textbf{90} & \textbf{120} & \textbf{150} & \textbf{60} & \textbf{90} & \textbf{120} & \textbf{150} \\
\midrule
\hline
DetectLRR & 0.5836 & 0.6164 & 0.6429 & 0.6492 & 0.7000 & 0.7501 & 0.7862 & 0.7904 & 0.6433 & 0.6814 & 0.6933 & 0.7146 \\
Lastde & \underline{0.5956} & \underline{0.6811} & \underline{0.7324} & \underline{0.7343} & 0.6184 & 0.6878 & 0.7273 & 0.7355 & 0.5829 & 0.6155 & 0.6301 & 0.6582 \\
\textbf{SpecDetect} & 0.5937 & 0.6562 & 0.7077 & 0.7081 & 0.7465 & 0.8268 & 0.8692 & 0.8802 & \underline{0.7133} & 0.7699 & 0.8136 & 0.8260 \\
\hline 
Fast-DetectGPT & 0.5631 & 0.6084 & 0.6506 & 0.6282 & 0.7834 & \underline{0.8411} & \underline{0.8668} & \underline{0.8824} & 0.6929 & \underline{0.7599} & \underline{0.7932} & \underline{0.8155} \\
Lastde++ & \textbf{0.5904} & \textbf{0.6579} & \textbf{0.7123} & \textbf{0.7052} & \textbf{0.7550} & \textbf{0.8578} & \textbf{0.9072} & \textbf{0.9219} & \textbf{0.6872} & \textbf{0.7981} & \textbf{0.8456} & \textbf{0.8713} \\
\textbf{SpecDetect++} & 0.6276 & 0.7032 & 0.7581 & 0.7566 & 0.7904 & 0.8814 & 0.9240 & 0.9365 & 0.7297 & 0.8212 & 0.8729 & 0.8880 \\
\bottomrule
\end{tabular}
}
\caption{Performance comparison across different methods with varying response lengths.}
\label{app_tab:text_length}
\end{table*}

\subsection{Impact of Contrast Sample Size}
\label{app:sample_size}

We present the detailed results across three datasets for varying numbers of contrast samples in Table~\ref{app_tab:sample_size}. Consistent with the analysis in the main body, these results demonstrate the robust effectiveness and superior sample efficiency of our method across all tested sample sizes.

\begin{table*}
\centering
\resizebox{\linewidth}{!}{
\begin{tabular}{l|ccccc|ccccc|ccccc}
\toprule
\textbf{Datasets} & \multicolumn{5}{c|}{\textbf{XSum}} & \multicolumn{5}{c|}{\textbf{WritingPrompts}} & \multicolumn{5}{c}{\textbf{Reddit}} \\
\textbf{Sample sizes} & \textbf{5} & \textbf{10} & \textbf{20} & \textbf{50} & \textbf{100} & \textbf{5} & \textbf{10} & \textbf{20} & \textbf{50} & \textbf{100} & \textbf{5} & \textbf{10} & \textbf{20} & \textbf{50} & \textbf{100} \\
\midrule
\hline
DNA-GPT & 0.7225 & 0.7225 & 0.7700 & 0.7350 & 0.7500 & \textbf{0.9675} & 0.9275 & 0.9475 & 0.9475 & 0.9550 & 0.8538 & 0.8800 & 0.8712 & 0.8800 & 0.8850 \\
Fast-DetectGPT & 0.6829 & 0.6705 & 0.6773 & 0.6870 & 0.6908 & 0.8874 & 0.9061 & 0.9156 & 0.9196 & 0.9228 & 0.8733 & 0.8850 & 0.9028 & 0.9118 & 0.9116 \\
Lastde++ & \underline{0.7935} & \underline{0.7924} & \underline{0.8004} & \underline{0.8167} & \underline{0.8196} & \underline{0.9552} & \underline{0.9624} & \underline{0.9694} & \underline{0.9723} & \underline{0.9751} & \underline{0.9427} & \underline{0.9515} & \underline{0.9643} & \underline{0.9702} & \underline{0.9699} \\
\textbf{SpecDetect++} & \textbf{0.8168} & \textbf{0.8317} & \textbf{0.8392} & \textbf{0.8516} & \textbf{0.8520} & \textbf{0.9593} & \textbf{0.9716} & \textbf{0.9737} & \textbf{0.9773} & \textbf{0.9797} & \textbf{0.9427} & \textbf{0.9516} & \textbf{0.9675} & \textbf{0.9709} & \textbf{0.9682} \\

\bottomrule
\end{tabular}
}
\caption{Performance comparison of different models with varying sample sizes.}
\label{app_tab:sample_size}
\end{table*}

\subsection{Generalization to Decoding Strategies.}
\label{app:decoding}
\begin{figure}[t]
    \centering
    \subfigure[Average AUC performance across three datasets under different top-k sampling.]{
        \includegraphics[width=1.\linewidth]{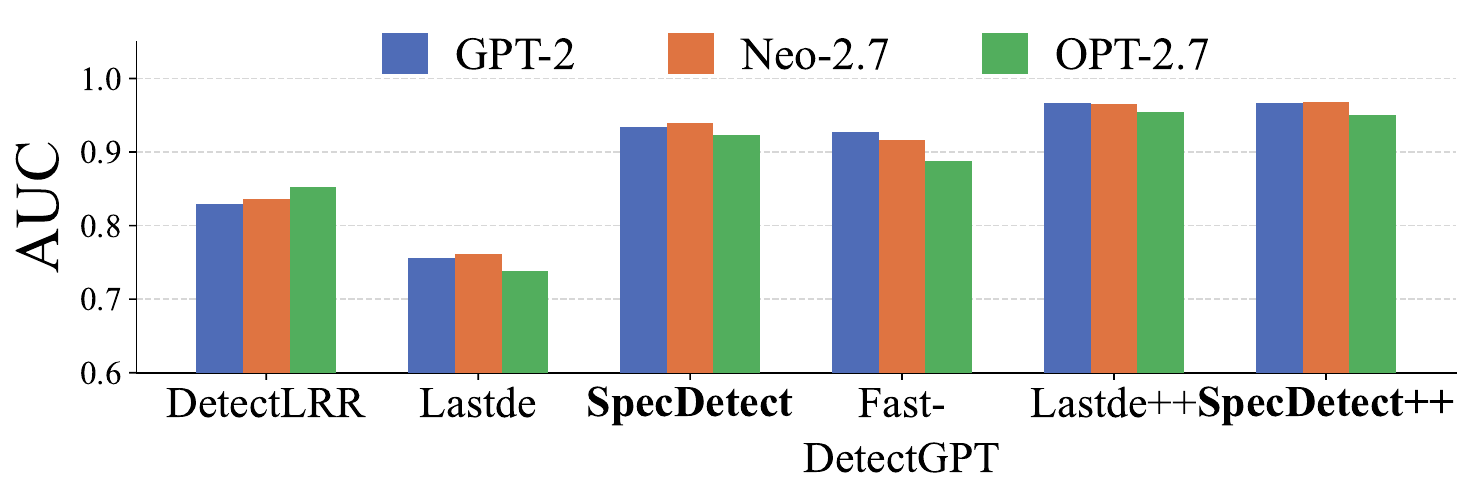}
    }
    \subfigure[Average AUC performance across three datasets under different top-p sampling.]{
        \includegraphics[width=1.\linewidth]{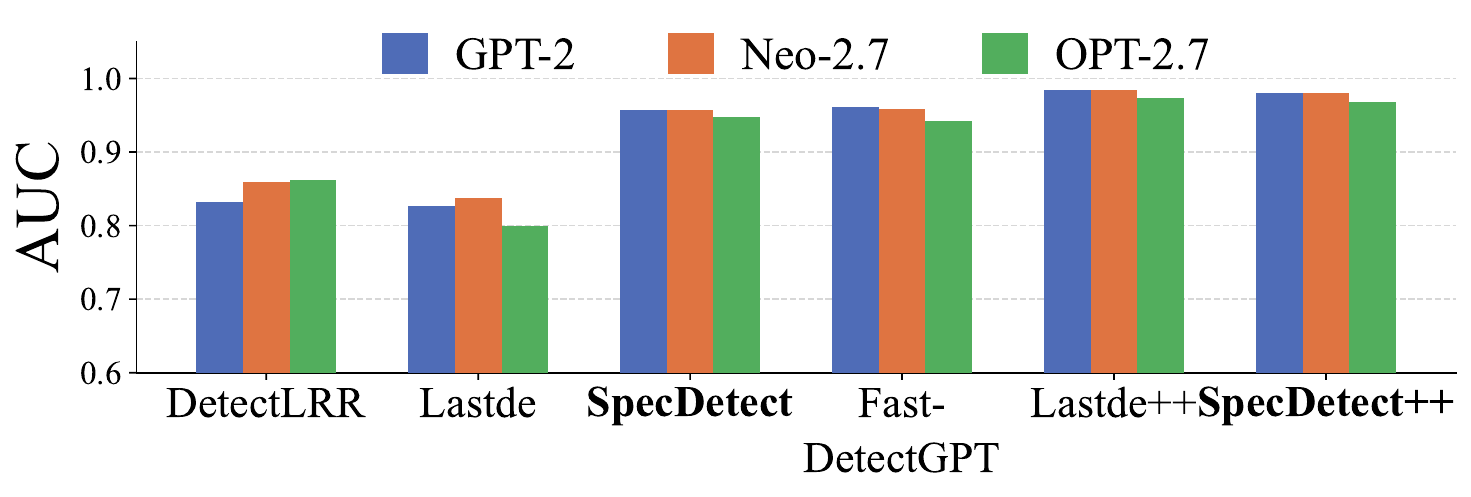}
    }
    \subfigure[Average AUC performance across three datasets under different temperature sampling.]{
        \includegraphics[width=1.\linewidth]{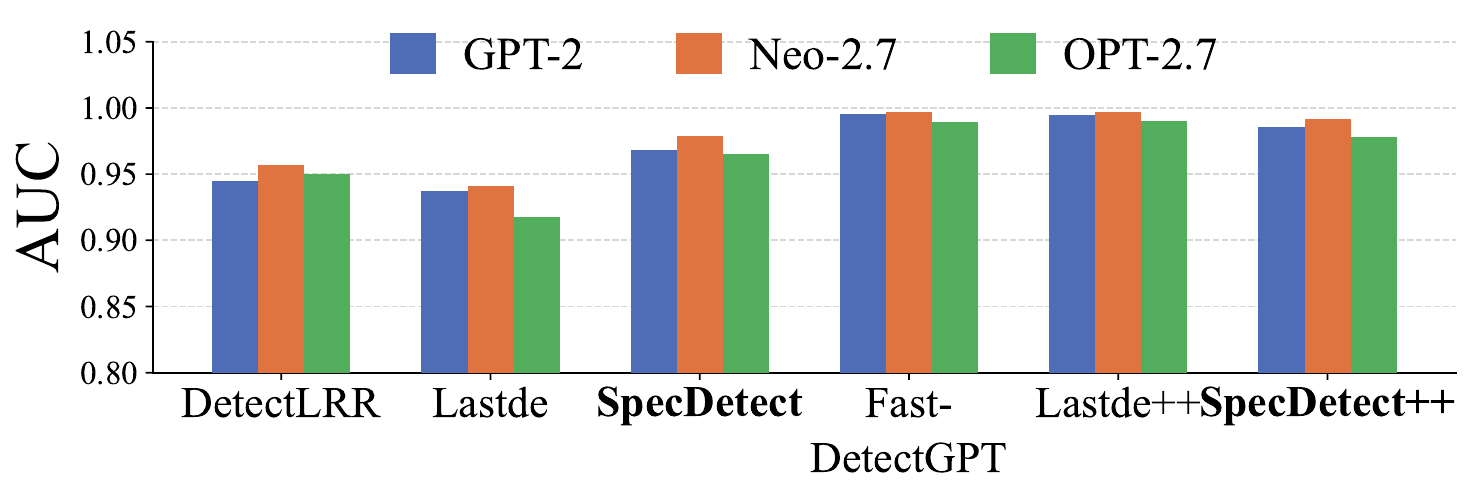}
    }
    \caption{Average AUC performance across three datasets under different decoding Strategies.}
    \label{fig:decoding_strategies}
\end{figure}

We evaluate the impact of different decoding strategies on detection performance. Figure~\ref{fig:decoding_strategies} presents the average AUC scores of various detectors across texts generated by three source models (GPT-2, Neo-2.7B, OPT-2.7B) using top-k \cite{fan2018hierarchical}, top-p \cite{holtzman2019curious}, and temperature sampling \cite{ackley1985learning}. Specifically, following the setup in previous work \cite{xu2024training}, we configure the parameters as: top-p = 0.96, top-k = 40, and temperature = 0.80.

The results reveal two key findings. First, as anticipated, the distribution-based methods, including our SpecDetect++, demonstrate state-of-the-art robustness and consistently form the outer, top-performing layer. A second, more striking finding is the robust performance of our base method, SpecDetect. Across all three decoding strategies and all source models, SpecDetect consistently and significantly outperforms all other sample-based methods. Its performance level is notably high for a single-pass detector, often approaching that of the distribution-based methods. This is a crucial result, as it indicates that our fundamental spectral energy feature is inherently more robust to variations in generation style than other statistical features, underscoring the fundamental nature of our frequency-domain approach.

\subsection{Performance of Alternative Metrics}
\label{app:alter_metrics}

In the main body, we identify energy-based features as the most effective for detection and select the global DFT Total Energy ($E_{DFT}$) based on the principle of parsimony. To validate this choice and provide a direct comparison of alternative metrics, we conduct an ablation study using our simple, sample-based \textbf{SpecDetect} framework as a consistent testbed. We evaluate the performance when the core $E_{DFT}$ feature is replaced with other energy-based metrics, including the local Total STFT Energy ($E_{STFT}$), the Mean Spectral Flux ($\overline{F}_{spec}$), and simple combinations (sum and product) of all three.

The results, presented in Table~\ref{tab:alternative_metrics}, confirm our initial hypothesis. While the individual energy-based metrics ($E_{DFT}$, $E_{STFT}$, and $\overline{F}_{spec}$) all demonstrate strong performance, their results are very similar, which is consistent with the high correlation we identified in our main analysis. Notably, attempting to combine these highly correlated features through simple summation or multiplication leads to a degradation in performance. This is likely because such combinations do not introduce new information and may amplify noise, reinforcing our decision to select a single, robust feature.

Furthermore, while the STFT-based metrics ($E_{STFT}$ and $\overline{F}_{spec}$) are effective, their calculation is dependent on the choice of STFT hyperparameters (e.g., window size and hop length). In contrast, the global $E_{DFT}$ is entirely hyperparameter-free, making it a more robust and generalizable choice. Therefore, this analysis confirms that the DFT Total Energy provides the best combination of high performance, simplicity, and robustness, validating its selection as the core feature for SpecDetect.

\begin{table*}[h]
\centering
\resizebox{\linewidth}{!}{
\begin{tabular}{lccccccccccccc}
\toprule
\textbf{Metric Used} & \textbf{GPT-2} & \textbf{Neo-2.7} & \textbf{OPT-2.7} & \textbf{LLaMA-13} & \textbf{LLaMA2-13} & \textbf{LLaMA3-8} & \textbf{OPT-13} & \textbf{BLOOM-7.1} & \textbf{Phi-4} & \textbf{Qwen3-8} & \textbf{Claude-3} & \textbf{GPT4-Turbo} & \textbf{Avg.}\\
\midrule
\textit{w/} $E_{DFT}$ & \textbf{0.9078} & \textbf{0.9171} & \textbf{0.9118} & \textbf{0.8051} & \textbf{0.8090} & \textbf{0.9761} & 0.9005 & \textbf{0.8781} & \textbf{0.7773} & \textbf{0.9783} & \textbf{0.9833} & \textbf{0.8054} & \textbf{0.8875} \\
\textit{w/} $E_{STFT}$ & 0.8977 & 0.9052 & 0.9054 & 0.7982 & 0.7998 & 0.9695 & 0.8967 & 0.8752 & \underline{0.7685} & 0.9569 & 0.9822 & 0.7993 & 0.8796 \\
\textit{w/} $\overline{F}_{spec}$ & 0.8827 & 0.8919 & 0.8863 & 0.7825 & 0.7815 & 0.9655 & 0.8806 & 0.8586 & 0.7345 & 0.9400 & 0.9738 & 0.7838 & 0.8635 \\
\textit{w/} Sum & \underline{0.9037} & 0.9114 & \underline{0.9107} & \underline{0.8023} & \underline{0.8045} & 0.9737 & \textbf{0.9015} & \underline{0.8771} & \underline{0.7685} & \underline{0.9589} & \textbf{0.9833} & \underline{0.8025} & \underline{0.8832} \\
\textit{w/} Product & 0.9028 & \underline{0.9115} & 0.9100 & 0.8016 & 0.8031 & \underline{0.9748} & \underline{0.9008} & 0.8763 & 0.7652 & 0.9581 & \underline{0.9829} & 0.8021 & 0.8824 \\

\bottomrule
\end{tabular}
}
\caption{Detection AUC of SpecDetect with alternative energy-based features. Results are averaged across three datasets in the black-box scenario. Best and second-best results are highlighted in \textbf{bold} and \underline{underline}, respectively.}
\label{tab:alternative_metrics}
\end{table*}
\end{document}